\documentclass[10pt,twocolumn,letterpaper]{article}
\usepackage[pagenumbers]{cvpr}
\usepackage[T1]{fontenc}
\usepackage{amsmath,amsfonts,booktabs,multirow,graphicx,array,colortbl}
\usepackage{nicefrac,microtype,makecell,float}
\usepackage{cuted,caption,subcaption}
\definecolor{cvprblue}{rgb}{0.21,0.49,0.74}
\usepackage[pagebackref,breaklinks,colorlinks,allcolors=cvprblue]{hyperref}
\title{Reimagine Video Dynamics: A Unified Framework for Video Editing\\with Editable Spatiotemporal Tokens}
\author{%
Yu Yuan$^{1,2}$\thanks{Work done during an internship at Adobe.}\quad
Yawen Lu$^{1}$\quad Guoxian Song$^{1}$\quad Kevin Duarte$^{1}$\\[4pt]
Ratheesh Kalarot$^{1}$\quad Di Chang$^{1}$\quad Xijun Wang$^{2}$\quad Stanley H. Chan$^{2}$\\[6pt]
$^{1}$Adobe Foundation Model\qquad $^{2}$Purdue University
}
\date{}
\hypersetup{
  pdftitle={Reimagine Video Dynamics: A Unified Framework for Video Editing with Editable Spatiotemporal Tokens},
  pdfauthor={Yu Yuan, Yawen Lu, Guoxian Song, Kevin Duarte, Ratheesh Kalarot, Di Chang, Xijun Wang, Stanley H. Chan}
}

\begin{document}
\maketitle
\begin{strip}
    \centering
    \includegraphics[width=\linewidth]{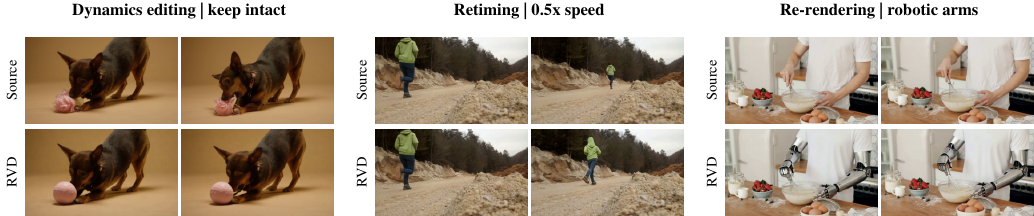}
    \captionof{figure}{RVD uses one renderer for three edits. Transforming the dynamics tokens changes the event, resampling them changes its pace, and reusing them preserves motion across appearance changes.}
    \label{fig:teaser}
\end{strip}

\begin{abstract}
Most video editing methods can change how a scene looks but offer limited control over how it moves. We introduce \textbf{Reimagine Video Dynamics (RVD)}, a unified framework for editing actions, motion pace, and appearance with reusable dynamics tokens and a shared renderer. The tokens capture motion from the source video and retain an explicit temporal axis. A language-guided editor transforms them to change an action or event. Resampling them changes the pace of motion, while keeping them fixed allows the renderer to reproduce the motion with a new appearance. We train the renderer to reconstruct source videos. We pretrain the editor on motion from real videos, then fine-tune it on counterfactual pairs. Across 450 dynamics edits, RVD achieves the highest instruction-following score and an 80.0\% pass rate. The same renderer supports retiming and motion reuse without additional task-specific training.
\par\smallskip\noindent Project page: \url{https://yuyuanspace.com/RVD/}.
\end{abstract}

\section{Introduction}
\label{sec:intro}
Modern video editing models can convincingly modify a video's appearance, from a runner's clothing to the background~\citep{Jiang_2025_VACE, Cheng_2024_InsV2V, Bai_2025_Ditto, Yang_2026_OmniVideo2, Yatim_2024_STDF, Yang_2025_VideoGrain, Chen_2026_VINO}. However, they often have limited control over the video's \emph{dynamics}, such as turning a run into a walk or making the runner stop before an obstacle. Such edits require changing motion while preserving the surrounding scene. Supporting these changes alongside temporal control and appearance re-rendering calls for a shared, editable representation of video dynamics.

Our central idea is to \textbf{learn a reusable dynamics interface for video editing}. We transform the source representation to change an action, resample it to adjust the pace, or keep it fixed to change appearance. A single renderer converts the result of each operation into video.

We introduce \textbf{Reimagine Video Dynamics (RVD)}, which consists of a dynamics encoder, a shared renderer, and a language-guided token editor. Through video reconstruction, the encoder and renderer learn compact tokens that retain a spatial grid and a temporal axis. The editor can then modify these tokens directly. The first frame and caption specify the scene's appearance.

For action editing, the source tokens tell the editor how motion unfolds in the original clip. For retiming, an index map specifies which source tokens to use at each output step. Learning semantic changes in this compact space lets the editor reuse the renderer for video synthesis.

To learn how an event could unfold differently, we construct counterfactual video pairs in which the scene retains its identity while an action, state, or interaction changes. Each pair includes an edit instruction. We first pretrain the editor on motion from real videos, then use the pairs to teach it how to transform source dynamics into target dynamics. Retiming and appearance changes use the trained renderer directly.

We compare complete systems on dynamics editing and study the shared representation through reconstruction, fixed-condition comparisons, motion reuse, and retiming. We make three contributions.
\begin{enumerate}
    \item We introduce a unified video-editing framework built on compact, editable spatiotemporal dynamics tokens and a shared renderer. The representation supports learned semantic transformations, direct temporal resampling, and reuse across appearance changes.
    \item We train a language-guided editor to transform source dynamics in token space through motion pretraining and fine-tuning on counterfactual pairs. The shared renderer converts the edited tokens into video.
    \item RVD supports training-free retiming and motion reuse, and achieves higher edit success than Ditto, OmniVideo2, and VINO. Controlled comparisons measure how the tokens affect reconstruction, motion retention, and edited events.
\end{enumerate}

\section{Related Work}
\label{sec:related}

\subsection{Video Editing Models}
Recent advances in generative modeling~\citep{Ho_2020_DDPM, Song_2021_Scorebased, Lipman_2023_FlowMatching, Rombach_2022_LDM, Peebles_2023_DiT} and large-scale video generation~\citep{Sora, Gao_2025_Seedance, Yang_2024_Cogvideox, Kong_2024_Hunyuanvideo, wan2025} have enabled a broad family of video editing models~\citep{Cheng_2024_InsV2V, Bai_2025_Ditto, Mai_2026_EasyV2V, Jiang_2025_VACE, Yang_2026_OmniVideo2, Chen_2026_VINO, Yatim_2024_STDF, Yang_2025_VideoGrain}. Most video editors focus on appearance. Fewer methods change how an event unfolds. Existing methods that edit a source video's dynamics often rely on explicit motion signals such as poses or sparse trajectories~\citep{Song_2025_XUniMotion, Tu_2024_MotionEditor, Mou_2024_ReVideo, Burgert_2025_MotionV2V}, specialize in object or interaction deletion~\citep{Motamed_2026_VOID}, plan trajectories at inference time~\citep{Yuan_2026_OptiWorld}, or leave dynamics control implicit in a pretrained generator~\citep{Kulikov_2026_DynaEdit}. Our focus is a shared, language-guided interface for editing source-video dynamics across common event families, with temporal control also accessible through direct token resampling.

\subsection{Representations for Video Dynamics}
Prior work separates appearance and motion in models such as MoCoGAN~\citep{Tulyakov_2018_MoCoGAN} for generation and VidStyleODE~\citep{Ali_2023_VidStyleODE} for video editing. RVD builds on this principle to learn instruction-conditioned source-to-target dynamics transformations from counterfactual video pairs, alongside direct temporal resampling. Existing video representations provide different starting points for this task.
\emph{Reconstruction-oriented representations}, such as video variational autoencoder (VAE)~\citep{wan2025, Yang_2024_Cogvideox} latents, preserve appearance, texture, and motion jointly for faithful decoding, making visual context and temporal evolution difficult to separate.
\emph{Semantic representations}, including the DINO family~\citep{Caron_2021_DINO, Oquab_2023_DINOv2, Simeoni_2025_DINOv3} and representation autoencoders (RAEs)~\citep{Zheng_2025_RAE}, emphasize object- and scene-level structure. Their training objectives place less emphasis on temporal evolution.
\emph{Predictive video representations}, such as the V-JEPA family~\citep{Bardes_2024_VJEPA, Assran_2025_VJEPA2, Murlabadia_2026_vjepa2_1}, instead learn spatiotemporal features by predicting latent video structure.
We use features from a frozen V-JEPA 2 encoder to learn our compact dynamics token.

\subsection{Physics-Aware Video Generation and Editing}
Physics-aware generation can be grouped by where physics enters the pipeline: generation followed by physical simulation~\citep{Xie_2024_Physgaussian, Zhang_2024_Physdreamer}, physical simulation followed by generative rendering~\citep{Liu_2024_Physgen, Savantaira_2024_Motioncraft, Xie_2025_Physanimator}, and generation guided by learned physical or motion priors~\citep{Li_2024_GenerativeImageDynamics, Chefer_2025_VideoJam, Xue_2025_PhyT2V, Yang_2025_VLIPP, Wang_2025_Wisa, Yuan_2026_NewtonGen} or inference-time planning and rewards~\citep{Yuan_2026_OptiWorld, Yuan_2026_WMReward}. Precise dynamics editing of real videos remains underexplored.

\section{Renderer: Learning Renderable Dynamics Tokens}
\label{sec:method}
We organize video rendering around two complementary inputs:
\begin{equation}
\mathbf{V}
=
\underbrace{\mathbf{C}}_{\text{visual context}}
+
\underbrace{\mathbf{D}}_{\text{dynamics}},
\label{eq:video_decomposition}
\end{equation}
where $\mathbf{C}$ specifies how the scene looks and $\mathbf{D}$ specifies how it evolves. The ``$+$'' denotes composition by the renderer. Changing $\mathbf{C}$ should preserve motion across appearances, while changing $\mathbf{D}$ should alter how the event unfolds within the chosen scene. We test both operations below.

\subsection{Renderer and Training}
\begin{figure*}[!t]
    \centering
    \includegraphics[width=\linewidth]{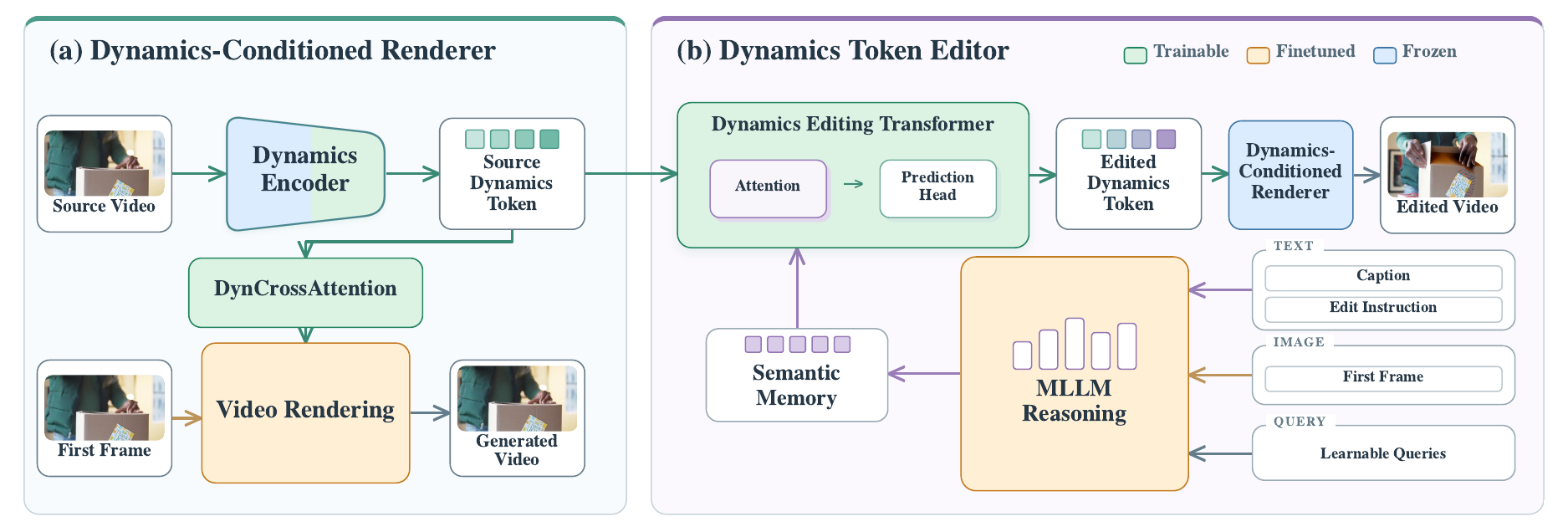}
    \caption{\textbf{RVD architecture.} \textbf{(a) Renderer:} Self-supervised reconstruction learns compact dynamics tokens and their rendering pathway. \textbf{(b) Editor:} A language-guided editor modifies the dynamics token, which is then rendered into a video with the target dynamics.}
    \label{fig:method}
\end{figure*}

\textbf{Learning Dynamics by Reconstruction.}
As shown in Fig.~\ref{fig:method}, the dynamics encoder extracts $\mathbf{D}$ from the source video. The first frame and an optional caption provide the visual context $\mathbf{C}=(\mathbf I_0,c)$. The renderer $R_\theta$ combines these inputs to reconstruct the source video,
\begin{equation}
\widehat{\mathbf V}=R_\theta(\mathbf I_0,c,\mathbf D).
\label{eq:renderer_overview}
\end{equation}
Here $\widehat{\mathbf V}$ denotes the reconstructed video.
Reconstruction jointly trains the dynamics representation and the adapters that pass it to the renderer.

\label{sec:renderer}
\label{sec:impl}
\textbf{Dynamics encoder.}
A frozen V-JEPA 2 ViT-L encoder~\citep{Assran_2025_VJEPA2} extracts spatiotemporal features from the source video. For an 81-frame clip, it produces 41 time steps on a $16\times16$ grid with 1024 channels. Modeling these high-dimensional features directly is costly and can be unstable~\citep{Nilaksh_2026_LatentWorldModels}. They may also retain appearance cues unrelated to motion. We use a learnable bottleneck to filter and compress the features. It reduces the temporal dimension from 41 to 21 steps. An overlapping $3\times3$ stride-2 convolution reduces the spatial grid to $8\times8$, and an MLP reduces the channel dimension to 128.
The resulting token contains about $64\times$ fewer values than the original V-JEPA 2 features and about $564\times$ fewer than the raw video.

\textbf{Dynamics-conditioned renderer.}
We build on Wan2.1-I2V-14B-480P~\citep{wan2025} and introduce a separate branch for dynamics tokens. We pass the first frame and caption through Wan's original conditioning paths and inject the dynamics token through DynCrossAttention adapters in the first 30 transformer blocks. To align the dynamics token with Wan features, we map the two spatial grids to the same continuous coordinate system and apply shared row and column rotary positional embeddings (2D-RoPE)~\citep{Su_2024_RoPE} to the queries and keys. This provides an explicit spatial correspondence between the dynamics token and the video latent.

We further constrain the dynamics pathway to emphasize motion while limiting changes to visual context. At each frame, we subtract the spatial mean of the adapter residual to reduce global appearance leakage. We also cap the root mean square (RMS) magnitude of the residual relative to the original Wan features. Finally, a diffusion-time gate gives the dynamics token stronger influence early in generation, when the overall motion is formed, and gradually reduces its influence later, when the model focuses more on visual details.

\begin{figure*}[!t]
\centering
\includegraphics[width=\linewidth]{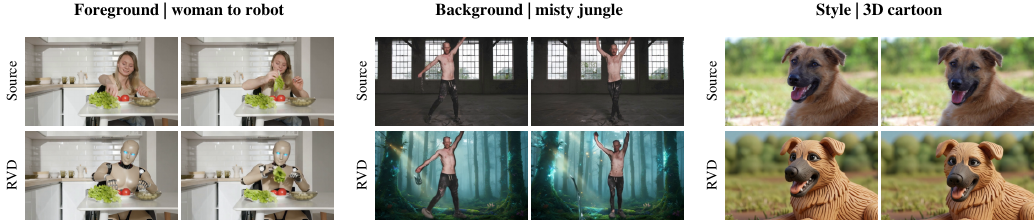}
\caption{\textbf{Changing scene appearance while preserving motion.} The renderer uses the same dynamics token with a different first frame and caption.}
\label{fig:rerender_examples}
\end{figure*}

\textbf{Renderer Training.} Following Wan2.1~\citep{wan2025}, let $\mathbf x_1$ be the clean video latent, $\mathbf x_0\sim\mathcal N(0,\mathbf I)$ be Gaussian noise, and $t$ be sampled from a logit-normal distribution. We form $\mathbf x_t=t\mathbf x_1+(1-t)\mathbf x_0$ and regress the rectified-flow velocity $\mathbf v_t=\mathbf x_1-\mathbf x_0$:
\begin{equation}
\mathcal L_{\mathrm{render}}=
\mathbb E_{\mathbf x_0,\mathbf x_1,t}
\left\|u_\theta(\mathbf x_t,t;\mathbf I_0,c,\mathbf D)-\mathbf v_t\right\|_2^2,
\label{eq:render_loss}
\end{equation}
where $u_\theta$ predicts the velocity conditioned on the first frame, caption, and dynamics token. Training uses about 169K filtered videos with diverse scenes, subjects, and motions. We apply caption dropout to encourage the renderer to rely on the dynamics token for motion reconstruction. We also perturb the dynamics token with random noise and masking to improve robustness and generalization. We first train the bottleneck and dynamics adapters from scratch, then fine-tune the Wan blocks with LoRA~\citep{Hu_2021_Lora}. We train the renderer for 6 days on 8 NVIDIA H200 GPUs. Appendix~\ref{app:renderer} provides additional architecture and training details.

\begin{figure*}[!t]
\centering
\includegraphics[width=\linewidth]{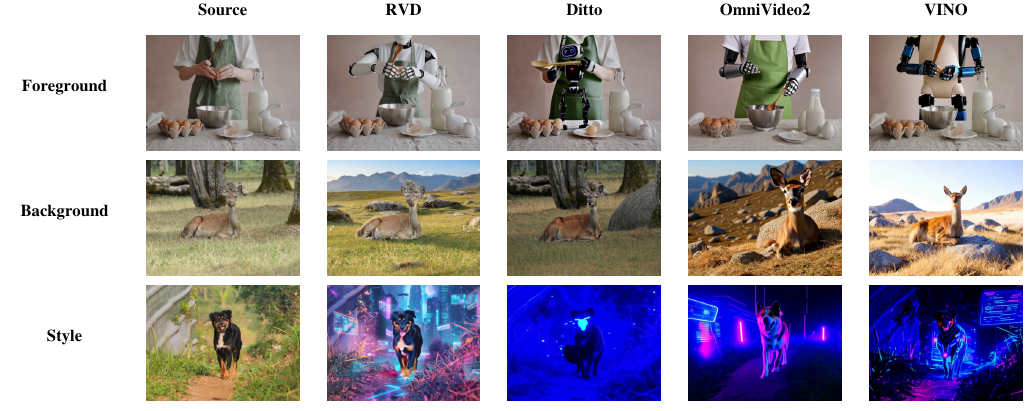}
\caption{\textbf{Qualitative comparison of video appearance re-rendering.}}
\label{fig:appearance_baseline_comparison}
\end{figure*}

\subsection{Rendering and Reusing Dynamics Tokens}
\label{sec:exp_renderer}

\textbf{Reconstruction.}
We first test whether the compact dynamics token preserves the motion information needed by the renderer. Given the first frame and extracted dynamics token, the renderer reproduces the source motion and reaches 19.261 dB PSNR. Removing the dynamics token sharply reduces motion consistency and lowers PSNR to 16.663 dB, as shown in Table~\ref{tab:renderer_arch_main}.

\textbf{Control through token interventions.}
We first fix the dynamics token $\mathbf D$ and change the first frame and caption. The foreground, background, or style changes while the output motion stays aligned with the source (Fig.~\ref{fig:rerender_examples} and Table~\ref{tab:flow_intervention}). Across 600 appearance edits, we compare the motion of each output with its source and with another source from the same edit type. We estimate optical flow at 12 matched times. Flow trajectory cosine averages 0.727 for matched pairs and 0.312 for mismatched pairs. The corresponding motion-profile correlations are 0.501 and 0.012, indicating that the output also tracks when motion in the source speeds up or slows down. We then fix the target first frame, caption, seed, and renderer and vary the dynamics input. The same scene now follows different event trajectories. Appendix~\ref{app:renderer_experiments} gives the protocols and results for both tests.

\textbf{Motion retention under appearance re-rendering.}
We test whether a fixed dynamics token can retain motion when the visual context changes. We compare with Ditto~\citep{Bai_2025_Ditto}, InsV2V~\citep{Cheng_2024_InsV2V}, OmniVideo2~\citep{Yang_2026_OmniVideo2}, STDF~\citep{Yatim_2024_STDF}, VideoGrain~\citep{Yang_2025_VideoGrain}, and VINO~\citep{Chen_2026_VINO}. Fig.~\ref{fig:rerender_examples} illustrates foreground, background, and style changes with the same dynamics token. Table~\ref{tab:renderer_selected} evaluates source--output V-JEPA similarity~\citep{Assran_2025_VJEPA2}, a proxy for motion retention, and VBench imaging quality~\citep{Huang_2024_VBench}. RVD achieves the highest JEPA similarity (0.9050) and an imaging quality score of 69.80. IF measures instruction following, and Pass gives the percentage of edits judged complete. The fixed-token flow intervention in Supplementary Table~\ref{tab:flow_intervention} provides an encoder-independent motion measurement. Fig.~\ref{fig:appearance_baseline_comparison} shows qualitative comparisons.

\begin{table*}[!t]
\centering
\caption{\textbf{Quantitative comparison of video appearance re-rendering.} Evaluation protocol: Appendix~\ref{app:renderer_benchmark}.}
\label{tab:renderer_selected}
\footnotesize
\begin{tabular*}{\linewidth}{@{\extracolsep{\fill}}lrrrrrrr@{}}
\toprule
Metric & \textbf{RVD} & Ditto & InsV2V & OmniVideo2 & STDF & VideoGrain & VINO \\
\midrule
IF $\uparrow$ & \textbf{4.438} & 3.973 & 1.615 & 3.712 & 2.085 & 2.260 & 4.345 \\
Pass (\%) $\uparrow$ & \textbf{86.8} & 74.5 & 10.5 & 63.5 & 26.0 & 27.2 & 81.3 \\
JEPA $\uparrow$ & \textbf{0.9050} & 0.8496 & 0.6933 & 0.8256 & 0.8688 & 0.8686 & 0.8706 \\
Imaging $\uparrow$ & 69.80 & 68.37 & 61.77 & \textbf{70.46} & 62.23 & 65.49 & 67.09 \\
\bottomrule
\end{tabular*}
\end{table*}

\subsection{Renderer Architecture Ablation}
\label{sec:renderer_ablation}
Table~\ref{tab:renderer_arch_main} studies three renderer design choices. First, removing dynamics conditioning causes the largest reconstruction drop, confirming that the token carries essential temporal information. Second, the $8\times8\times128$ bottleneck offers a balance between reconstruction fidelity and token count. The larger spatial grid improves fidelity at a substantially higher cost. Third, 2D RoPE and residual centering provide the clearest gains among the injection choices, while injecting into every block or removing the noise gate has a smaller effect. Appendix~\ref{app:renderer_ablation} describes the shared evaluation protocol.
\begin{table*}[!t]
\centering
\caption{\textbf{Renderer ablations across conditioning, bottleneck capacity, and dynamics injection.}}
\label{tab:renderer_arch_main}
\footnotesize
\begin{minipage}[t]{0.54\textwidth}
\centering
\textbf{(a) Conditioning and injection}\\[-2pt]
\begin{tabular*}{\linewidth}{@{\extracolsep{\fill}}llcc@{}}
\toprule
Design & Variant & PSNR $\uparrow$ & SSIM $\uparrow$ \\
\midrule
\multirow{3}{*}{Conditioning}
& Full renderer & 19.261 & 0.5761 \\
& No dynamics token & 16.663 & 0.5216 \\
& No caption & 18.630 & 0.5770 \\
\midrule
\multirow{5}{*}{Injection}
& Full renderer & 19.261 & 0.5761 \\
& No 2D RoPE & 18.222 & 0.5511 \\
& No centering & 18.427 & 0.5743 \\
& Inject all blocks & 19.093 & 0.5853 \\
& No noise gate & 19.054 & 0.5855 \\
\bottomrule
\end{tabular*}
\end{minipage}
\hfill
\begin{minipage}[t]{0.42\textwidth}
\centering
\textbf{(b) Bottleneck capacity}\\[-2pt]
\begin{tabular*}{\linewidth}{@{\extracolsep{\fill}}lcc@{}}
\toprule
Token shape & PSNR $\uparrow$ & SSIM $\uparrow$ \\
\midrule
$4{\times}4{\times}128$ & 18.691 & 0.5723 \\
$8{\times}8{\times}64$ & 18.346 & 0.5818 \\
$8{\times}8{\times}128$ & 19.261 & 0.5761 \\
$8{\times}8{\times}256$ & 17.963 & 0.5591 \\
$8{\times}8{\times}512$ & 18.166 & 0.5615 \\
$8{\times}8{\times}1024$ & 18.467 & 0.5659 \\
$16{\times}16{\times}128$ & 19.593 & 0.5960 \\
\bottomrule
\end{tabular*}
\end{minipage}
\end{table*}

\section{Editor: Editing the Learned Dynamics Space}
\label{sec:editing_system}
The same renderer decodes tokens after temporal resampling or semantic editing. To change an event, the editor takes source dynamics $\mathbf D$, an instruction $q$, source and target captions $c_s,c_t$, and a target first frame $I_t$. It predicts the edited token
\begin{equation}
\widehat{\mathbf D}=T_\xi(\mathbf D,q,c_s,c_t,I_t).
\label{eq:editor_token}
\end{equation}
The renderer substitutes $\widehat{\mathbf D}$ for $\mathbf D$ in Eq.~\ref{eq:renderer_overview} to generate the target video.

\begin{figure*}[!t]
\centering
\includegraphics[width=\linewidth]{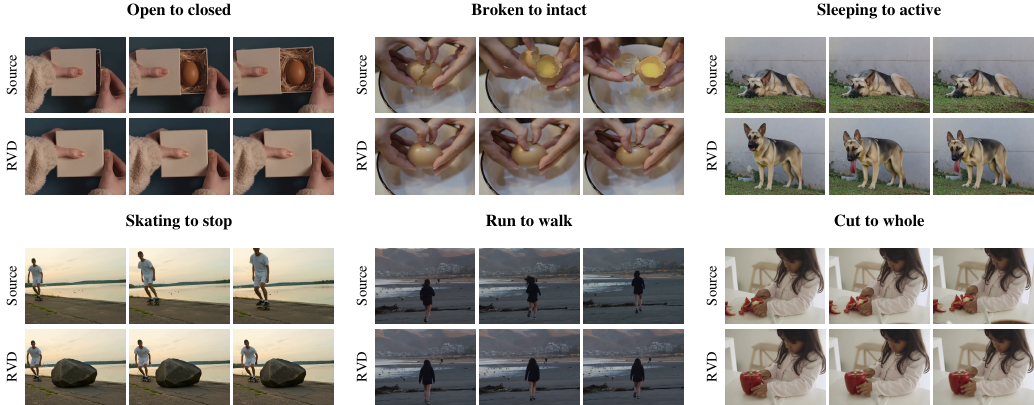}
\caption{\textbf{Video dynamics editing with RVD.}}
\label{fig:editor_examples}
\end{figure*}

\subsection{Token Structure and Direct Temporal Control}
\label{sec:token_structure}
\label{sec:topology}

We first examine which variations survive compression, then resample the temporal axis to control motion pace.

\begin{figure}[!t]
\centering
\includegraphics[width=\linewidth]{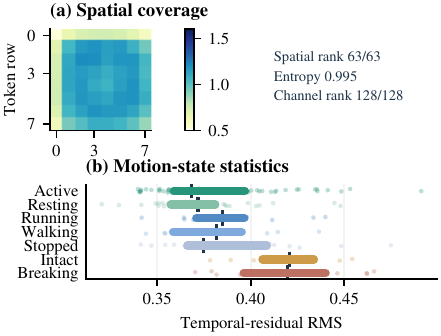}
\caption{\textbf{Dynamics-token analysis.} \textbf{(a)} Temporal variation spans the spatial grid. \textbf{(b)} Motion states exhibit different but overlapping token distributions.}
\label{fig:token_structure}
\end{figure}

\textbf{Structure of the dynamics space.}
We analyze post-bottleneck tokens from 248 real videos. Temporal variation spans the full $8\times8$ spatial grid, with normalized coverage entropy $0.995$ and maximum spatial rank $63/63$ (Fig.~\ref{fig:token_structure}). The distribution of residual energy shifts across source-motion states, with overlap between groups. Channel covariance also retains full rank ($128/128$). Appendix~\ref{app:token_analysis} defines these measures and reports additional statistics.

\begin{figure*}[!t]
\centering
\includegraphics[width=\linewidth]{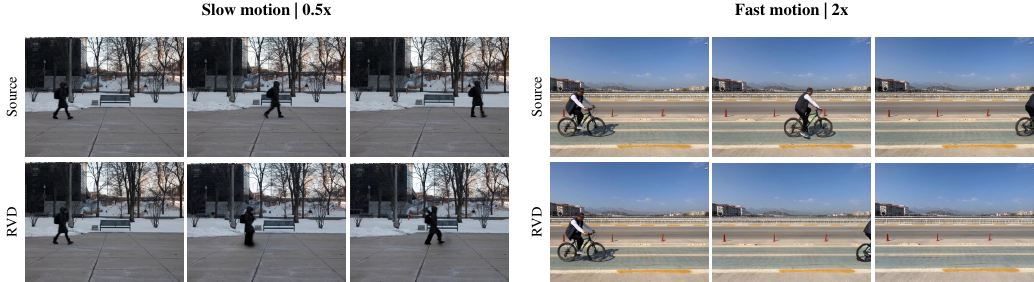}
\caption{\textbf{Training-free retiming in dynamics-token space.}
Resampling the dynamics token along its temporal axis directly slows down or speeds up the rendered motion without additional training.}
\label{fig:retiming_examples}
\end{figure*}

\textbf{Direct manipulation through retiming.}
To retime a video, we specify which source token positions condition each output step. This index map sets the token schedule exactly, and the renderer generates motion from that schedule. Stretching the token sequence slows the motion, while subsampling it speeds the motion up. Both operations preserve visual content in Fig.~\ref{fig:retiming_examples} and use the trained renderer directly. The speed labels give the token resampling factors. Appendix~\ref{app:retime} describes the implementation and measures the effect on rendered motion.

\textbf{Quantitative retiming results.}
For each of 311 videos, we render the original token and its two resampled versions using the same first frame, caption, and noise seed. For a requested factor $r$, we compare optical flow in each output against flow between the source frames specified by the map $\phi(i)=ri$. At $0.5\times$, the reference covers the full output. At $2\times$, the source frames span the first half of the output, after which the token sequence is zero-padded. Resampling improves flow alignment at both speeds (Table~\ref{tab:retiming_quantitative}). At $0.5\times$, flow cosine rises from 0.2302 to 0.5986. At $2\times$, it rises from 0.1985 to 0.5845 over the first half. Appendix~\ref{app:retime_quantitative} gives the flow metrics and sampling protocol.

\begin{table*}[!t]
\centering
\caption{\textbf{Retiming improves motion correspondence with the requested source time map.} We compare retimed and unchanged outputs on 311 videos. Intervals give paired 95\% bootstrap confidence intervals for the difference.}
\label{tab:retiming_quantitative}
\small
\begin{tabular}{llrrrr}
\toprule
Target / evaluated interval & Metric & Unchanged $1\times$ & Retimed RVD & Paired difference [95\% CI] & RVD wins \\
\midrule
$0.5\times$ / full clip & Flow cosine $\uparrow$ & 0.2302 & \textbf{0.5986} & $+0.3684$ [0.3322, 0.4038] & 86.5\% \\
 & Normalized error $\downarrow$ & 1.5452 & \textbf{0.8410} & $-0.7042$ [$-0.7616$, $-0.6514$] & 96.8\% \\
$2\times$ / first half & Flow cosine $\uparrow$ & 0.1985 & \textbf{0.5845} & $+0.3860$ [0.3547, 0.4162] & 92.6\% \\
 & Normalized error $\downarrow$ & 1.0278 & \textbf{0.7725} & $-0.2553$ [$-0.2802$, $-0.2284$] & 93.9\% \\
\bottomrule
\end{tabular}
\end{table*}

\begin{figure}[!t]
\centering
\includegraphics[width=\linewidth]{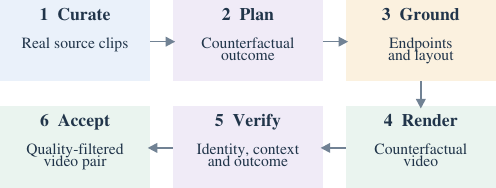}
    \caption{\textbf{Counterfactual data pipeline.} }
    \label{fig:data_pipeline}
\end{figure}

\subsection{Video Dynamics Editor}
\label{sec:editor}
The editor transforms the source dynamics token according to a natural-language instruction. A multimodal encoder interprets the requested change, and a transformer applies it in dynamics space (Fig.~\ref{fig:method}(b)). Fig.~\ref{fig:editor_detail} shows the architecture in detail.

\textbf{Semantic memory.} We adapt Qwen3-VL-4B-Instruct~\citep{Bai_2025_Qwen3VL} to provide semantic guidance for dynamics editing. It receives three input types: text containing the source and target captions and edit instruction, the target first frame, and eight learnable queries that gather and compress task-relevant information. Qwen jointly reasons over these inputs, and the final query states form compact semantic tokens that guide the dynamics editor. The Qwen backbone and vision encoder remain frozen, while rank-4 LoRA and the learnable queries provide lightweight adaptation.

\textbf{Dynamics editing transformer.}
The editor uses tokens from across the source clip to model how its motion unfolds. It first flattens the source token into a spatiotemporal sequence. In each transformer block, self-attention connects positions across space and time, and cross-attention reads the semantic tokens. A prediction head maps the edited features back to the original token shape for rendering.

At inference, an MLLM~\citep{Bai_2025_Qwen3VL} infers the target-video caption from the source video and edit instruction. If the requested dynamics require a different initial pose or state, Qwen-Image-Edit~\citep{Wu_2025_QwenImage} also produces a target first frame; otherwise, we reuse the source first frame. The edited dynamics token, target caption, and target first frame are then passed to the trained renderer to generate a video with the requested dynamics.

\begin{figure*}[!t]
\centering
\includegraphics[width=\linewidth]{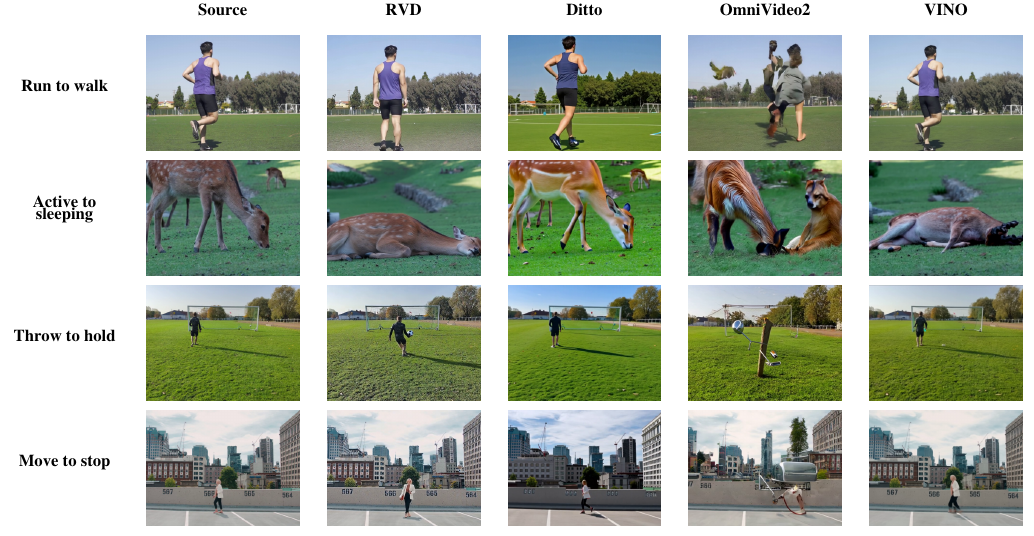}
\caption{\textbf{Qualitative comparison of video dynamics editing.}}
\label{fig:dynamics_baseline_comparison}

\end{figure*}

\begin{table*}[!t]
\centering
\footnotesize
\setlength{\tabcolsep}{4pt}
\caption{\textbf{Video dynamics editing on all 450 candidate cases.}}
\label{tab:editor_selected}
\begin{tabular}{lrrrr}
\toprule
\multirow{2}{*}{Method} & \multicolumn{2}{c}{Edit success} & \multicolumn{2}{c}{Output-internal consistency} \\
\cmidrule(lr){2-3} \cmidrule(lr){4-5}
 & IF $\uparrow$ & Pass (\%) $\uparrow$ & Background stability $\uparrow$ & Temporal flicker consistency $\uparrow$ \\
\midrule
\rowcolor{gray!10}
RVD (ours) & \textbf{4.144} & \textbf{80.0} & 0.9607 & \textbf{0.9908} \\
Ditto~\citep{Bai_2025_Ditto} & 2.582 & 38.2 & 0.9555 & 0.9862 \\
OmniVideo2~\citep{Yang_2026_OmniVideo2} & 2.098 & 25.6 & 0.9534 & 0.9781 \\
VINO~\citep{Chen_2026_VINO} & 3.842 & 71.3 & \textbf{0.9644} & 0.9870 \\
\bottomrule
\end{tabular}
\end{table*}

\subsection{Editor Training: Counterfactual Data and Two-Stage Strategy}
\label{sec:data}

\textbf{Counterfactual video pair construction.}
Training the dynamics editor requires pairs of videos that share the same visual context but exhibit different dynamics. Such pairs are difficult to collect from real videos, so we build the counterfactual data pipeline shown in Fig.~\ref{fig:data_pipeline}. Starting from a real video, we first identify its motion category and use an MLLM~\citep{Bai_2025_Qwen3VL} to plan a counterfactual outcome. We then edit the target layout with Qwen-Image-Edit~\citep{Wu_2025_QwenImage} and synthesize the target video with VACE~\citep{Jiang_2025_VACE}. An MLLM-based quality check removes samples with inconsistent identity, background, or target motion. Finally, each accepted pair is reused in the reverse direction by swapping source and target and reversing the instruction. Of 11,702 generated candidates, 9,964 pass the quality check (85.1\%), and 9,946 pairs remain after decoding and cache checks. Using both directions gives 19,892 training examples spanning common dynamics families, including activity, locomotion, object-state changes, and interactions. For generation details and statistics, see Appendix~\ref{app:counterfactual_data}.

\textbf{Two-stage training.}
We first construct 169K pseudo dynamics pairs from real videos: a static source repeats the first frame, and the target is the corresponding real-video dynamics token. This pretraining exposes the editor to realistic motion beyond the counterfactual pairs. We then fine-tune on 20K directed counterfactual examples to learn source-to-target transformations in the dynamics space while reusing the renderer. Retiming and appearance reuse need no separate training. Editor training takes approximately 5 days on 8 NVIDIA H200 GPUs; Appendix~\ref{app:training_data_comparison} compares reported training-data scopes.

\textbf{Token-space Loss.}
We train the editor in dynamics-token space. Let $\widehat{X}$ and $X^{*}$ denote the predicted and target tokens. We decompose each into its temporal mean and residual,
\begin{equation}
X=\mu(X)+A(X)
\end{equation}
where $\mu(X)$ captures the temporal mean component and $A(X)$ captures temporal variation.

We use $\mathcal L_{\mathrm{dc}}$ to match the temporal means and $\mathcal L_{\mathrm{ac}}$ to match the temporal residuals. A cosine loss $\mathcal L_{\mathrm{cos}}$ further aligns the direction of temporal changes. We also match residual variance with $\mathcal L_{\mathrm{var}}$ to preserve motion magnitude, and temporal frequency spectra with $\mathcal L_{\mathrm{fft}}$ to preserve motion pace and rhythm. The complete objective is
\begin{equation}
\mathcal L_{\mathrm{edit}}
=
\lambda_{\mathrm{dc}}\mathcal L_{\mathrm{dc}}
+
\lambda_{\mathrm{ac}}\mathcal L_{\mathrm{ac}}
+
\lambda_{\mathrm{cos}}\mathcal L_{\mathrm{cos}}
+
\lambda_{\mathrm{var}}\mathcal L_{\mathrm{var}}
+
\lambda_{\mathrm{fft}}\mathcal L_{\mathrm{fft}}.
\label{eq:loss_v2}
\end{equation}
For the loss definitions and optimization settings, see Appendix~\ref{app:editor_training}.

\textbf{Training ablations.}
Table~\ref{tab:editor_training_ablation}(a) compares two editor initializations on the intact-to-break task using 118 training pairs and eight validation pairs. Both use the same token representation and frozen renderer. At the same update budget, pretraining improves temporal-residual prediction. Panel (b) shows how removing each token-space loss affects the corresponding prediction error.

\begin{table}[!t]
\centering
\caption{\textbf{Dynamics-editor ablations.} \textbf{(a)} Fixed-budget initialization diagnostic (100 updates). \textbf{(b)} Token-space loss ablations report changes from their full-loss reference.}
\label{tab:editor_training_ablation}

\begin{minipage}[t]{\linewidth}
\centering
\footnotesize
\textbf{(a) Two-stage training}\\[-1pt]

\begin{tabular}{lcc}
\toprule
Training strategy & AC-cos $\uparrow$ & AC MSE $\downarrow$ \\
\midrule
W/o Two-stage & 0.2629 & 0.6977 \\
Two-stage
& \textbf{0.2892} & \textbf{0.6352} \\
\bottomrule
\end{tabular}
\end{minipage}
\par\medskip
\begin{minipage}[t]{\linewidth}
\centering
\footnotesize
\textbf{(b) Dynamics-aware loss}\\[-1pt]

\begin{tabular}{llc}
\toprule
Removed & Metric & Change \\
\midrule
$\mathcal L_{\mathrm{cos}}$
& AC-cos $\uparrow$
& $-0.0134$ \\

\multirow{2}{*}{$\mathcal L_{\mathrm{var}}+\mathcal L_{\mathrm{fft}}$}
& Variance error $\downarrow$
& $+0.2079$ \\

&
Spectrum error $\downarrow$
& $+0.3276$ \\
\bottomrule
\end{tabular}
\end{minipage}

\end{table}

\subsection{Dynamics Editing Results}
\label{sec:exp_editor}
\textbf{Qualitative comparison.}
Fig.~\ref{fig:editor_examples} shows edits to outcomes, activities, locomotion, and object states. A state replacement can begin in the requested state. A transition requires the change to unfold on screen. In Fig.~\ref{fig:dynamics_baseline_comparison}, Ditto~\citep{Bai_2025_Ditto}, OmniVideo2~\citep{Yang_2026_OmniVideo2}, and VINO~\citep{Chen_2026_VINO} often retain the source motion, while RVD changes it as requested.

\textbf{Benchmark results.}
Across all 450 candidate scenes, RVD achieves the highest instruction-following score, pass rate, and temporal flicker consistency (Table~\ref{tab:editor_selected}). Each system receives the same source video and instruction through its native interface and produces one output per case without reranking. Qwen3-VL-8B-Instruct scores instruction following and completion from 12 uniformly sampled source/output frames using deterministic decoding. The VBench metrics~\citep{Huang_2024_VBench} measure stability within each output. The supplement gives the full protocol.

\textbf{What the dynamics pathway adds.}
With the target frame, caption, renderer, and sampling fixed, edited tokens yield the highest within-output appearance consistency (Supplementary Table~\ref{tab:editor_necessity}). The same tokens also support direct retiming and motion reuse across appearances with no additional editor training.

\textbf{Human evaluation.}
Six experts scored anonymous candidates in two panels of three. Each panel assigned one joint score per dimension. Averaged across panels, RVD ranks first on all three dimensions (Table~\ref{tab:human_main}). Appendix~\ref{app:human_questionnaire} gives the protocol and full results.

\begin{table}[H]
\centering
\caption{\textbf{Human evaluation on 34 cases per method.} Scores range from 1 to 3 (higher is better). C: event completion; P: temporal-process correctness; R: unrelated-content preservation.}
\label{tab:human_main}
\small
\begin{tabular}{lccc}
\toprule
Method & C $\uparrow$ & P $\uparrow$ & R $\uparrow$ \\
\midrule
OmniVideo2 & 1.015 & 1.044 & 1.015 \\
Ditto & 1.206 & 1.265 & 1.353 \\
VINO & 1.559 & 1.544 & 1.676 \\
RVD & \textbf{2.029} & \textbf{1.941} & \textbf{1.941} \\
\bottomrule
\end{tabular}
\end{table}

\textbf{Reusing the learned interface.}
Semantic editing, temporal resampling, and appearance changes use the same renderer (Supplementary Table~\ref{tab:token_operations}). After real-video reconstruction and static-to-real pretraining, we train the editor on 19,892 directed examples. The other two operations require no further RVD training. Appendix~\ref{app:training_data_comparison} compares reported data budgets.

\section{Conclusion}
\label{sec:conclusion}
RVD uses a shared dynamics representation to edit actions, retime motion, and change appearance. A language-guided editor transforms source tokens to change how an event unfolds. Resampling those tokens adjusts motion pace, and reusing them carries source motion into a new visual context. In the human evaluation, RVD ranks first on event completion, temporal-process correctness, and content preservation. All three operations use one renderer, with no additional task-specific training for retiming or appearance changes.

\clearpage
\section*{LLM Usage Statement}
Generative AI tools (Codex, Claude Code) assisted with language editing and figure preparation, and were used within the research pipeline to generate and filter counterfactual training pairs and to perform automated video evaluation. The authors designed these protocols, audited their outputs, verified the reported results, and take full responsibility for the paper.
{\small\bibliographystyle{ieeenat_fullname}\bibliography{refs}}
\clearpage
\begin{center}\Large\bfseries Appendix\end{center}
\appendix
\setcounter{figure}{0}
\setcounter{table}{0}
\setcounter{equation}{0}
\renewcommand{\theHfigure}{appendix.\arabic{figure}}
\renewcommand{\theHtable}{appendix.\arabic{table}}
\renewcommand{\theHequation}{appendix.\arabic{equation}}

\renewcommand{\thefigure}{S\arabic{figure}}
\renewcommand{\thetable}{S\arabic{table}}
\renewcommand{\theequation}{S\arabic{equation}}
\section{Renderer: Architecture, Training, Experiments, and Ablations}
\label{app:renderer}

\subsection{Renderer Architecture}
\begin{figure*}[t]
\centering
\includegraphics[width=\textwidth]{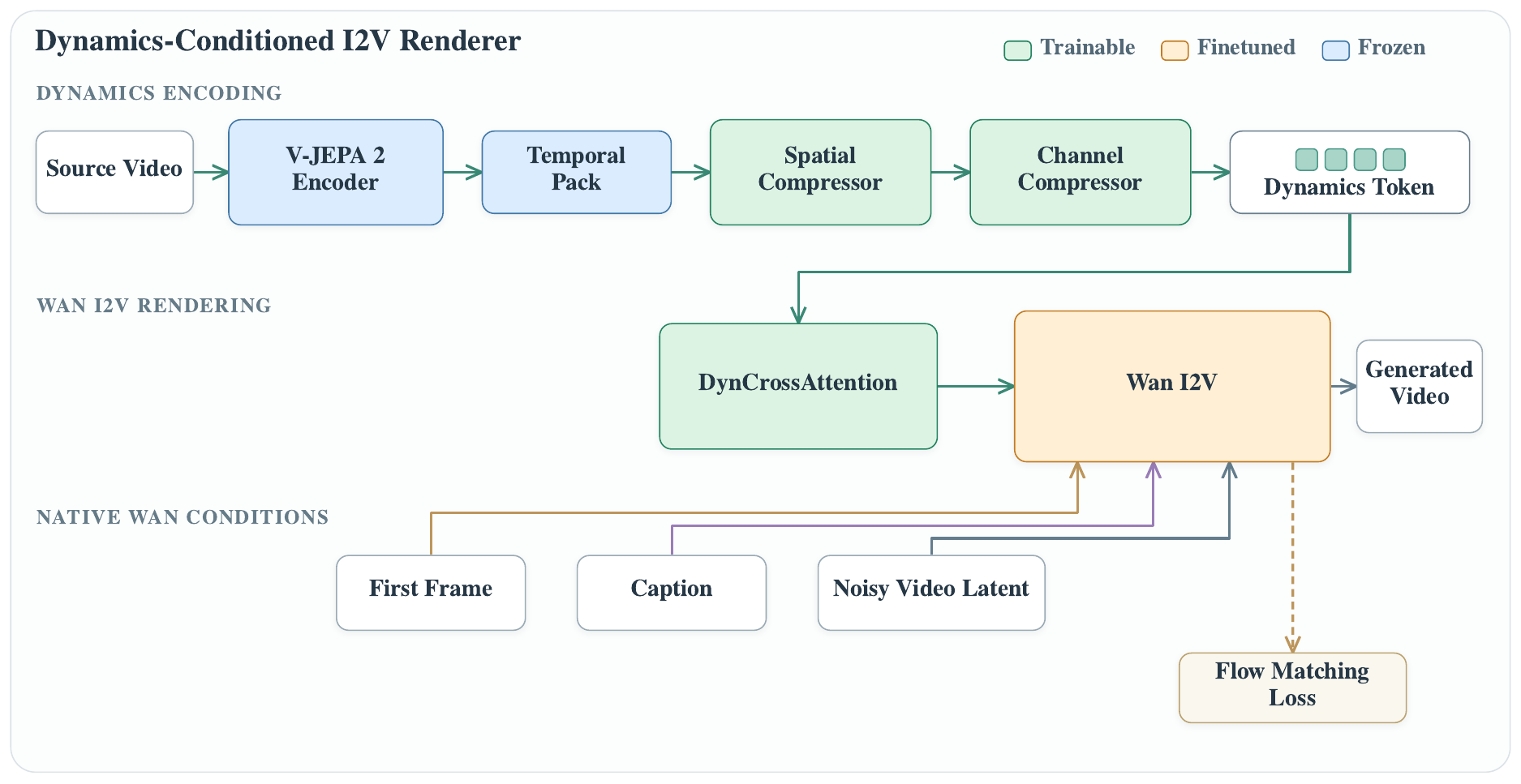}
\caption{\textbf{Detailed renderer architecture.} Frozen V-JEPA features pass through the trainable dynamics bottleneck. DynCrossAttention injects only the compressed dynamics token into the finetuned Wan I2V backbone; image, text, and noisy-latent conditions retain Wan's native paths.}
\label{fig:renderer_detail}
\end{figure*}

Figure~\ref{fig:renderer_detail} expands the renderer in Fig.~\ref{fig:method}(a). An 81-frame video first enters the frozen V-JEPA 2 ViT-L encoder, producing a $41\times16\times16\times1024$ feature volume. The learnable bottleneck normalizes these features, packs adjacent temporal positions, reduces the spatial grid with an overlapping $3\times3$ stride-2 convolution, and projects channels with an MLP. 

The target first frame, caption, and noisy video latent follow the native Wan2.1-I2V-14B-480P paths. Dynamics use a separate DynCrossAttention adapter in the first 30 of 40 Wan blocks. At each video time step, the adapter uses the Wan latent patches as queries and the corresponding $8\times8$ dynamics slice as keys and values. It aligns their spatial coordinates with shared 2D RoPE, computes same-time cross-attention, and adds the result to the Wan features as a bounded residual. This preserves Wan's native image and text conditioning while giving the renderer an explicit path for temporal evolution.

For latent frame $t$, let $H_t\in\mathbb R^{P\times d}$ denote Wan patch features and $D_t\in\mathbb R^{64\times128}$ the corresponding dynamics slice. The adapter computes
\begin{align}
Q_t&=\operatorname{RoPE}_{2D}(W_q\operatorname{LN}(H_t)),\\
K_t&=\operatorname{RoPE}_{2D}(W_k\operatorname{LN}(D_t)),\\
U_t&=W_o\operatorname{softmax}(Q_tK_t^\top/\sqrt d)W_v\operatorname{LN}(D_t).
\end{align}
Wan coordinates are continuously rescaled onto the $8\times8$ token grid before applying shared row and column RoPE. We center the adapter output, $\bar U_t=U_t-\operatorname{mean}_{p}U_t$, force $\bar U_0=0$, and update
\begin{equation}
H_t' = H_t + g(\sigma_t)\min\!\left(1,\frac{0.05\,\operatorname{RMS}(H_t)}{\operatorname{RMS}(\bar U_t)}\right)\bar U_t.
\end{equation}
Centering suppresses global appearance shifts, the RMS cap prevents the adapter from overwhelming Wan features, and the noise gate $g$ emphasizes dynamics early in denoising. Every Wan patch attends to all 64 dynamics locations from the same time step.

\paragraph{Trainable parameters.}
The renderer optimizes approximately 437.3M parameters: 332.5M in the bottleneck and DynCrossAttention adapters and 104.9M in rank-32 Wan LoRA modules. V-JEPA 2 and the 14B Wan backbone remain frozen.

\subsection{Renderer Training}
We train the renderer on approximately 169K internally collected real-world clips covering diverse subjects, environments, viewpoints, and everyday motions. We select clips with one dominant event, stable visual quality, and continuous motion. Each clip is standardized to 81 frames at 24 fps and serves as its own reconstruction target, so training requires only the videos.

We use the rectified-flow objective in Eq.~\ref{eq:render_loss}. Caption dropout encourages the renderer to use the dynamics token for motion. Perturbing or masking tokens during training prepares the renderer for the editor's predicted tokens. We first train the bottleneck and adapters, then enable rank-32 LoRA on Wan. Training takes six days on eight NVIDIA H200 GPUs.

\subsection{Rendering and Reusing Dynamics Tokens}
\label{app:renderer_experiments}
We expand on the three experiments in Section~\ref{sec:exp_renderer}. We test how well the token preserves motion during reconstruction, vary the visual context and dynamics separately, and compare appearance re-rendering with existing video editors.

\begin{figure*}[!t]
\centering
\includegraphics[width=\textwidth]{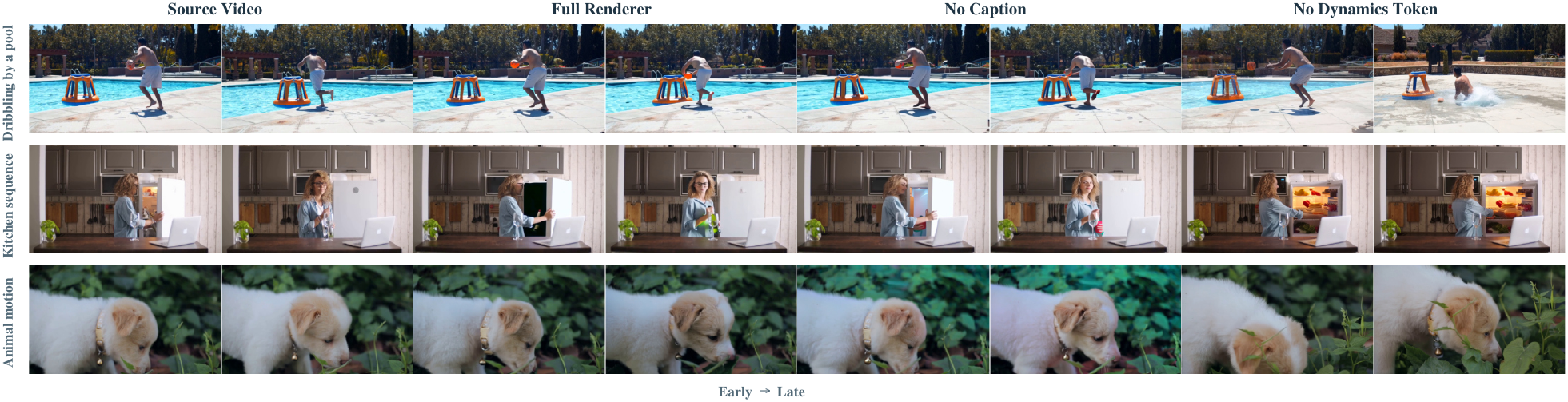}
\caption{\textbf{Reconstruction under controlled renderer conditions.} The full and caption-free renderers retain the source event, whereas removing the dynamics token changes or suppresses its evolution.}
\label{fig:renderer_condition_ablation}
\end{figure*}

\paragraph{Reconstruction.}
Given the first frame, caption, and compact token, the renderer reaches 19.261 dB PSNR and 0.5761 SSIM. Removing the token lowers both scores to 16.663 dB and 0.5216. Removing the caption gives 18.630 dB and 0.5770, a small increase in SSIM with a decrease in PSNR. In Fig.~\ref{fig:renderer_condition_ablation}, both the full and caption-free renderers follow the source event. Removing dynamics changes or suppresses its progression.

\paragraph{Control through token interventions.}
We test both directions with all other generation inputs held constant. In the first direction, we fix $\mathbf D$ and change only the first frame and caption. Figure~\ref{fig:rerender_examples} shows foreground, background, and style interventions under this setting. Table~\ref{tab:flow_intervention} reports the matched and mismatched comparisons.

We measure motion preservation at 12 matched times using Farneb\"ack flow at $192\times112$. Horizontal flow, vertical flow, and magnitude are pooled over a $4\times4$ grid to form a trajectory descriptor, while interval magnitudes form a motion-energy profile. The negative control pairs each output with another source from the same edit type.

\begin{figure*}[t]
\centering
\includegraphics[width=\textwidth]{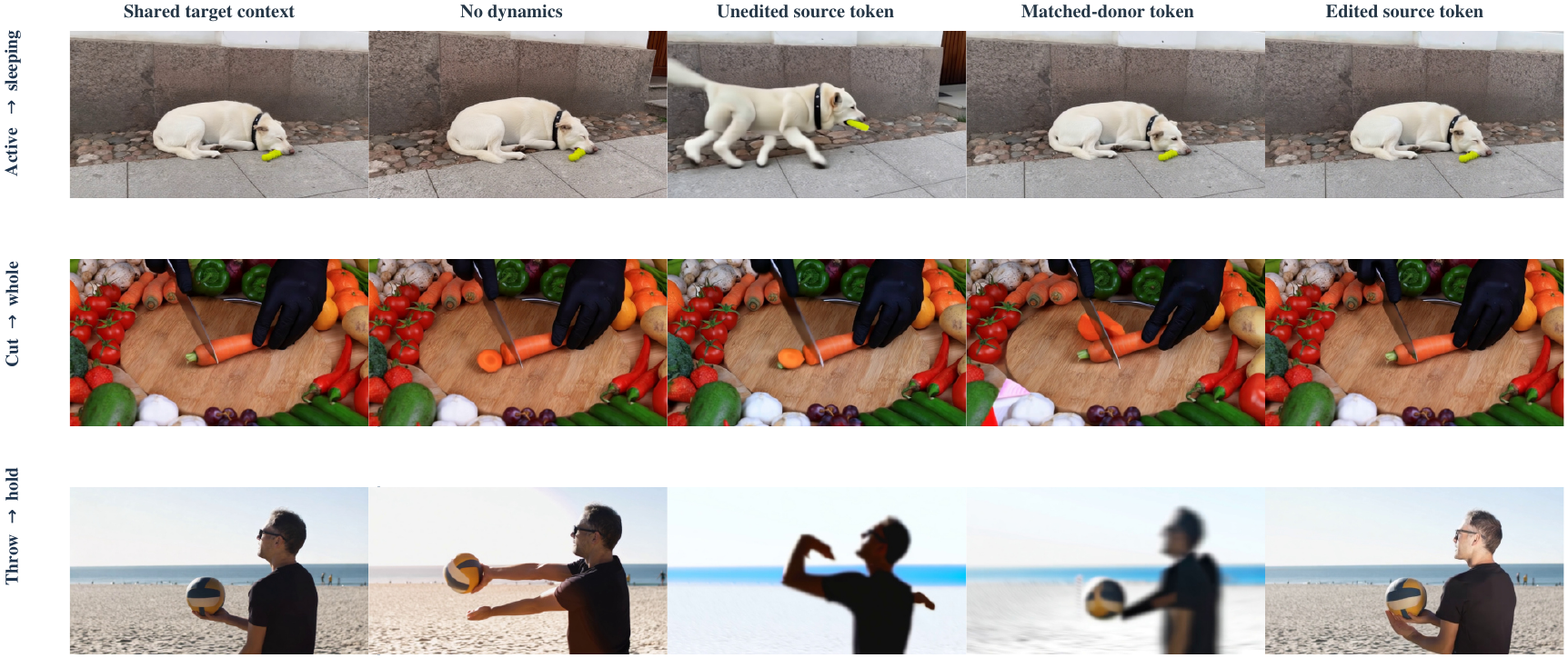}
\caption{\textbf{Editor contribution under fixed target context.} All conditions use the same target context, instruction, seed, and renderer; only the dynamics input changes.}
\label{fig:dynamics_switch}
\end{figure*}

In the reverse direction, we fix the target first frame, caption, edit instruction, seed, sampling schedule, and renderer, and change only the dynamics input. Figure~\ref{fig:dynamics_switch} compares no token, the unedited source token, a same-task donor token, and the edited source token. Under identical visual context, the displayed token interventions yield different motion progressions. The edited-token examples illustrate control over progression relative to an obstacle, travel distance, and scene framing (Figs.~\ref{fig:token_benefits} and~\ref{fig:token_benefits_state}). Quantitatively, edited source dynamics achieve the highest within-output DINO appearance consistency among the controlled conditions (Table~\ref{tab:editor_necessity}). This complementary intervention shows that changing dynamics alters temporal evolution without requiring a change to the visual context. Appendix~\ref{app:fixed_context_editor} provides additional controlled evaluations across edit families.

These interventions show how each input affects the rendered video. A fixed token can carry motion across appearance changes, although a token from another video may also introduce foreground cues from that video. Editing the source token keeps the target context fixed and gives the highest within-output DINO appearance consistency in Table~\ref{tab:editor_necessity}.

\begin{table*}[t]
\centering
\caption{Direct motion preservation when visual context changes under fixed $\mathbf D$. The mismatched control pairs each output with another source from the same edit type.}
\label{tab:flow_intervention}
\small
\setlength{\tabcolsep}{7pt}
\begin{tabular}{lrr}
\toprule
Pairing & Flow trajectory $\uparrow$ & Motion profile $\uparrow$ \\
\midrule
Matched source--output & \textbf{0.727} & \textbf{0.501} \\
Mismatched source--output & 0.312 & 0.012 \\
\bottomrule
\end{tabular}
\end{table*}

\begin{figure*}[t]
\centering
\includegraphics[width=\textwidth]{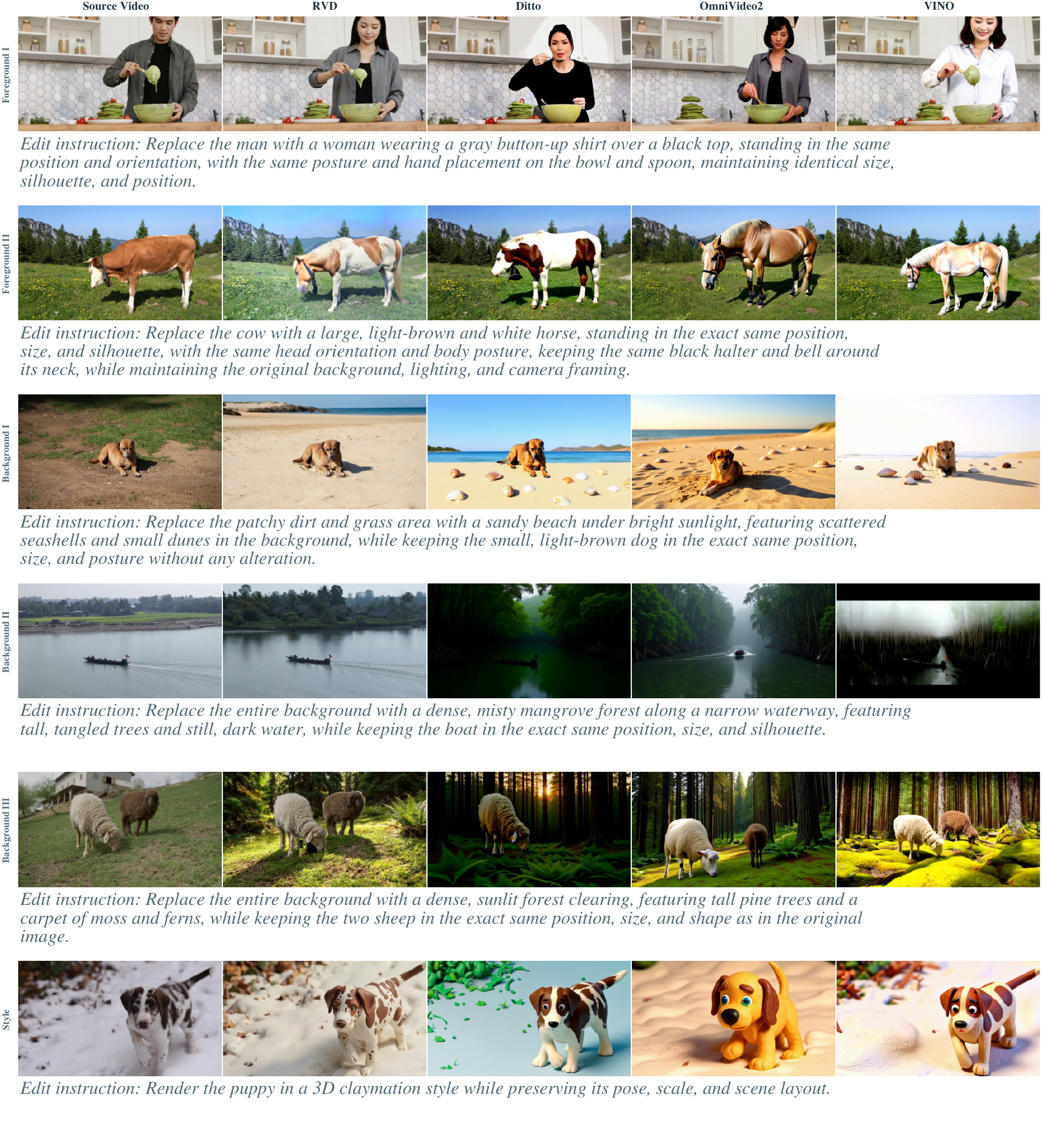}
\caption{\textbf{More renderer comparisons.} Six appearance edits compare RVD with Ditto~\citep{Bai_2025_Ditto}, OmniVideo2~\citep{Yang_2026_OmniVideo2}, and VINO~\citep{Chen_2026_VINO} at aligned relative time; each row reports its instruction.}
\label{fig:renderer_baseline_gallery}
\end{figure*}

\paragraph{Motion retention under appearance re-rendering.}
The comparison covers foreground, background, and style edits (Appendix~\ref{app:renderer_benchmark}). We evaluate motion retention and rendering quality using one output per case, without reranking. JEPA is source--output V-JEPA similarity, a motion-retention proxy that can also reflect visual semantics; Imaging is VBench image quality. Table~\ref{tab:renderer_selected} shows that RVD achieves the highest JEPA similarity (0.9050) and an Imaging score of 69.80. The independent flow intervention in Table~\ref{tab:flow_intervention} complements this feature-based measurement. Figure~\ref{fig:renderer_baseline_gallery} provides qualitative examples for all three appearance changes.

\subsection{Renderer Ablations}
\label{app:renderer_ablation}
In Table~\ref{tab:renderer_arch_main}, we vary one renderer component at a time. All variants use the same reconstruction split, preprocessing, first frame, caption policy, and PSNR/SSIM evaluation.

\paragraph{Conditioning.}
No dynamics sets the adapter input to zero; no caption removes text while retaining the first frame and token. The 2.598 dB PSNR loss without dynamics is the largest conditioning drop, demonstrating the contribution of dynamics conditioning to reconstruction fidelity.

\paragraph{Bottleneck capacity.}
We change only the post-bottleneck spatial grid or channel width and retrain the corresponding projection. The $8\times8\times128$ token provides the selected efficiency--fidelity tradeoff: $4\times4$ loses spatial detail, while $16\times16$ improves PSNR by only 0.332 dB at four times the spatial-token count. Increasing channels beyond 128 does not improve reconstruction.

\paragraph{Injection design.}
No 2D RoPE removes shared spatial coordinates; no centering keeps the adapter's spatial mean; inject-all-blocks extends adapters from 30 to 40 blocks; and no noise gate holds $g(\sigma_t)$ constant. Removing RoPE or centering lowers PSNR to 18.222 and 18.427 dB, respectively, giving the clearest evidence for spatial alignment and appearance-neutral residuals. Extending injection or removing the gate provides no PSNR gain over the selected design.

\section{Dynamics Token Analysis and Retiming}
\label{app:dynamics_analysis}

\subsection{Post-bottleneck Token Analysis}
\label{app:token_analysis}
We analyze the compressed representation used by both the renderer and the editor. Removing duplicate source-video IDs leaves 248 real clips. For each token $\mathbf D\in\mathbb R^{21\times8\times8\times128}$, we separate the temporal mean from its varying component,
\begin{equation}
\mu(\mathbf D)=\operatorname{mean}_t\mathbf D, \qquad
A(\mathbf D)=\mathbf D-\mu(\mathbf D).
\end{equation}

We use three complementary measurements. First, channel covariance treats all clip--time--location positions as observations; all 128 eigen-directions remain above $10^{-5}$ of the largest eigenvalue, giving rank $128/128$. Second, spatial motion energy $e_s=\mathbb E_{n,t,c}[A(\mathbf D)_{n,t,s,c}^2]$ covers the full $8\times8$ grid with normalized entropy 0.995, and centered spatial slices reach the maximum possible rank $63/63$. Third, Fig.~\ref{fig:token_structure}(b) plots each clip's residual RMS. Different source-motion groups shift the distribution, although they overlap because appearance, viewpoint, and motion magnitude also vary. After compression, temporal variation still spans channels and spatial locations. We use spatiotemporal self-attention to model this structure.

\subsection{Training-free Retiming}
\label{app:retime}
\begin{figure*}[t]
\centering
\includegraphics[width=\textwidth]{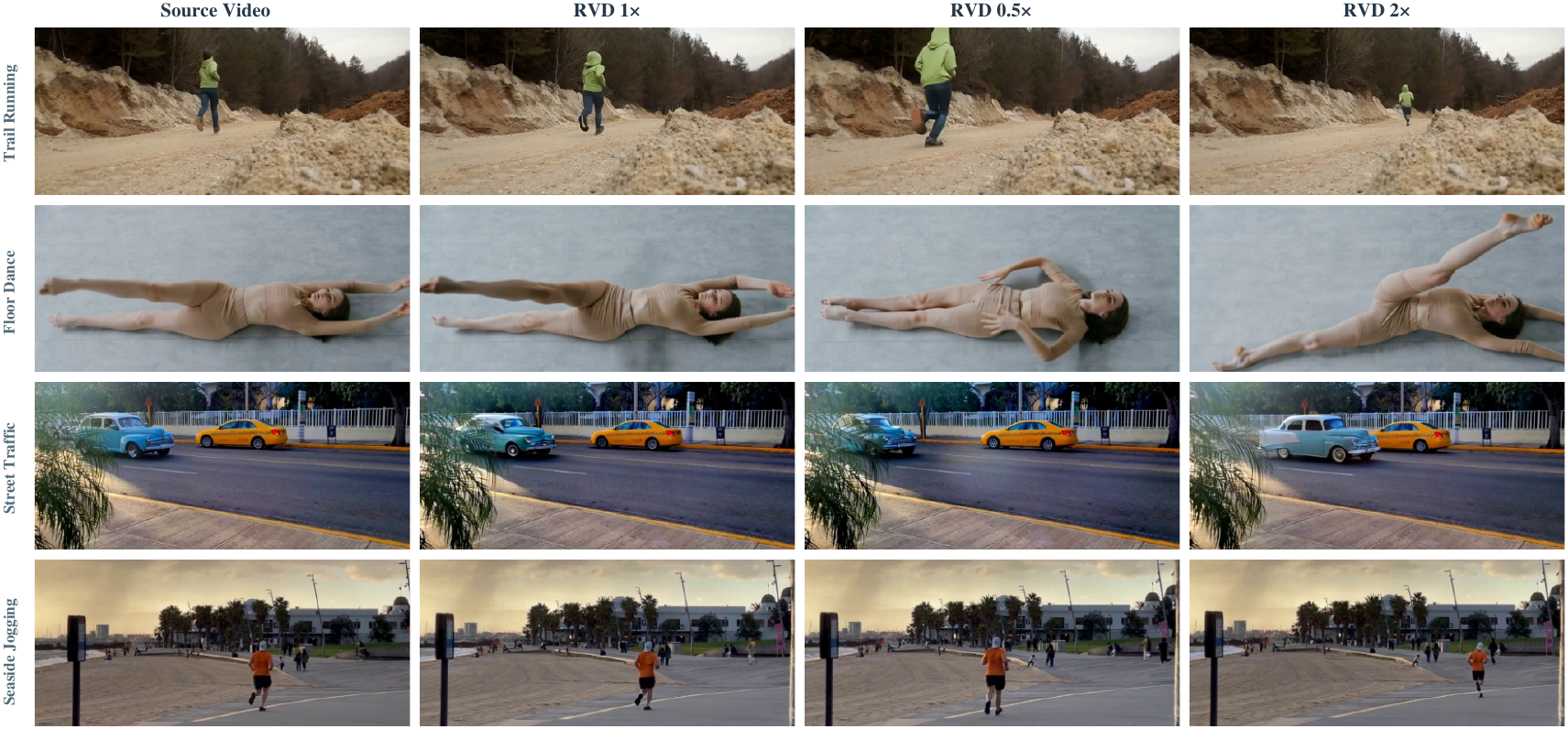}
\caption{\textbf{Additional training-free retiming results.} At the same normalized time, the $0.5\times$ output has progressed less and the $2\times$ output has progressed further; the $1\times$ reconstruction follows the source timing.}
\label{fig:retiming_appendix}
\end{figure*}

The explicit temporal axis permits direct speed control without training another model. For half-speed motion, nearest-neighbor temporal resampling reads the token at $\phi(t)=0.5t$ and a width-3 temporal smoother removes repeated-step discontinuities:
\begin{equation}
\texttt{slow}_{0.5\times}(\mathbf D)=\operatorname{smooth}_3\!\left(\operatorname{resample}(\mathbf D,\phi(t)=0.5t)\right).
\end{equation}
For double speed, we retain every second token through step 20 and set the unused future to zero:
\begin{equation}
\texttt{fast}_{2\times}(\mathbf D)= [\mathbf D_0,\mathbf D_2,\ldots,\mathbf D_{20};\mathbf 0_{10\text{ steps}}].
\end{equation}
Both operators preserve the token shape and use the frozen renderer. In the fast-motion examples, motion continues during the zero-padded tail. The $2\times$ label gives the resampling stride over the available source tokens. Figure~\ref{fig:retiming_appendix} compares four videos at matched normalized times. In each example, the $0.5\times$ output progresses less than the source, the $2\times$ output progresses further, and the reconstruction follows the source timing.

\subsection{Quantifying the Effect of Temporal Resampling}
\label{app:retime_quantitative}
We use all 311 source IDs in the existing retiming pool for which the source, $1\times$, $0.5\times$, and $2\times$ clips are available. All 1,244 files decode to 81 frames at 16 fps. File availability was the sole inclusion criterion. For each source, the three outputs share the first frame, caption, renderer, and noise seed.

\paragraph{Target time map and comparison.}
For output frame $i$ and requested factor $r$, we use source frame $\phi(i)=ri$ as the reference for both the retimed output and the unchanged $1\times$ control. At $0.5\times$, output indices $0,4,\ldots,80$ map to source indices $0,2,\ldots,40$. At $2\times$, indices $0,4,\ldots,40$ map to $0,8,\ldots,80$. We evaluate this first half because the zero-padded tail has no corresponding source trajectory. We use the specified frame indices directly, without post hoc alignment. The factors are defined in frame-index space.

\paragraph{Metrics.}
We resize frames to $192\times112$ and estimate consecutive sampled-frame Farneb\"ack flow with fixed parameters (pyramid scale 0.5, three levels, window 15, three iterations, polynomial size 5 and sigma 1.2). Flow cosine compares concatenated mean horizontal/vertical vectors over a $4\times4$ grid across all sampled intervals. Let $F$ and $F^*$ denote output and target-reference flow. The normalized flow error is
\begin{equation}
 E(F,F^*)=\frac{\sum w\,\|F-F^*\|_2}{\sum w\,\|F^*\|_2},\qquad w=\|F^*\|_2+10^{-3},
\end{equation}
where sums cover pixels and intervals. Target-motion weighting emphasizes moving regions without selecting cases. Each clip contributes equally to the aggregate. Confidence intervals use 5,000 paired case-bootstrap replicates with seed 20261116. Digitally resampled source clips yield cosine 1 and error 0 against themselves, as expected in this implementation check.

\paragraph{Interpretation and tail behavior.}
Both operators improve flow correspondence with the target time map relative to the unchanged token (Table~\ref{tab:retiming_quantitative}). Motion continues in frames 44--80 of the $2\times$ output, where the token is zero-padded. Mean flow magnitude in this interval is 0.637 pixels per sampled interval, compared with 0.628 for $1\times$. These measurements include camera motion and depend on flow estimation and spatial alignment.

\section{Editor: Architecture, Data, Training, and Experiments}
\label{app:data}

\subsection{Editor Architecture}
\begin{figure*}[t]
\centering
\includegraphics[width=\textwidth]{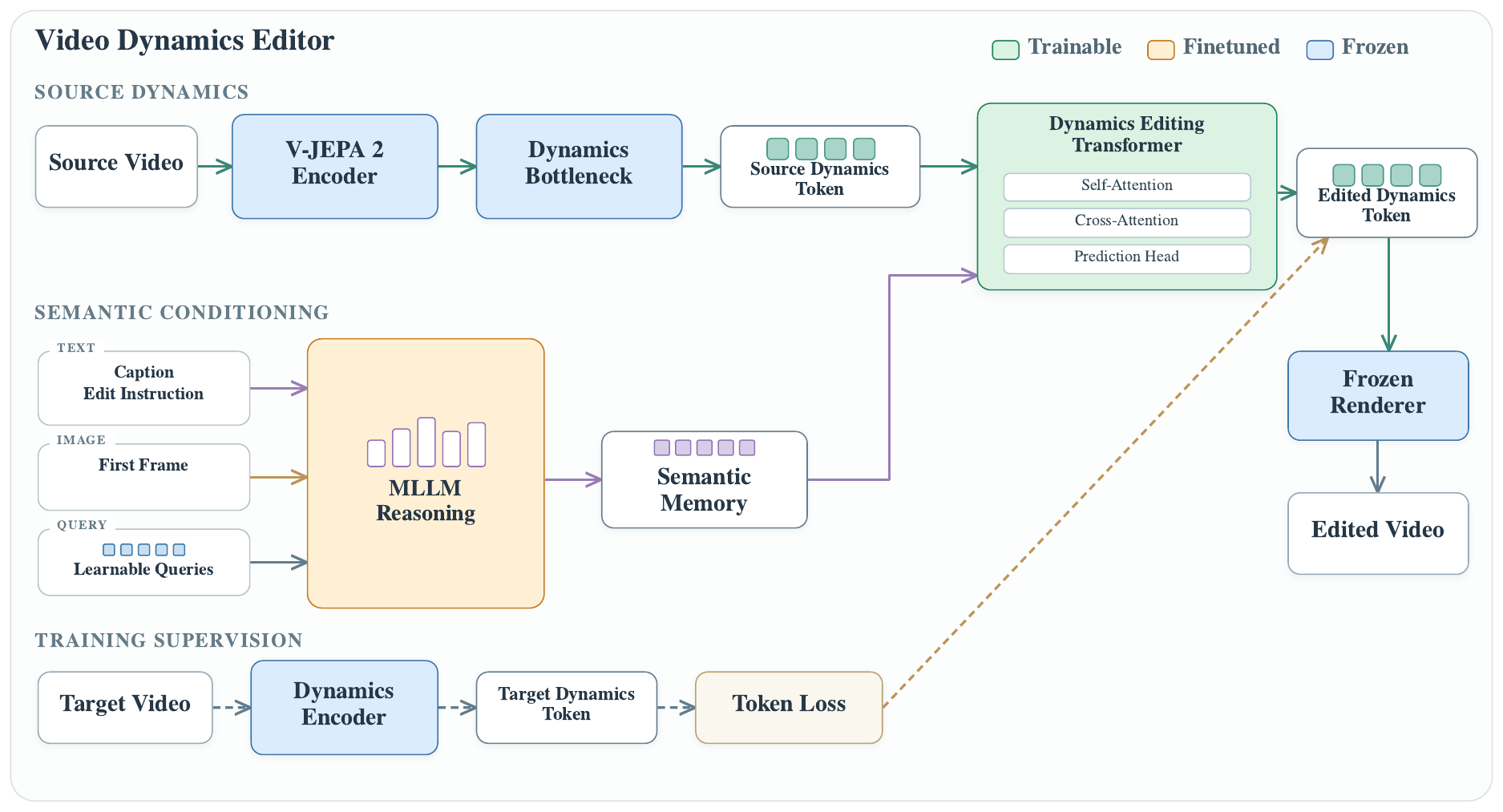}
\caption{\textbf{Detailed editor architecture.} Text, image, and learned queries enter the finetuned MLLM independently. Its semantic memory guides the trainable token editor through cross-attention and AdaLN; the target-token branch and token loss are training-only.}
\label{fig:editor_detail}
\end{figure*}

Figure~\ref{fig:editor_detail} shows the editor from Fig.~\ref{fig:method}(b) in more detail. The encoder maps the source video to a standardized $21\times8\times8\times128$ dynamics token. The semantic branch uses Qwen3-VL-4B-Instruct with rank-4 attention LoRA. It takes the source and target captions and edit instruction as text, processes the target first frame through its vision encoder, and uses eight learned queries to summarize the requested change. The final query states form the semantic memory.

The four-layer dynamics editing transformer has width 256 and four attention heads. Coordinate embeddings retain the token's time and $8\times8$ location. Self-attention models dependencies within source dynamics; cross-attention reads semantic memory; and AdaLN uses the pooled semantic state to modulate each block. A $256\!\rightarrow\!512\!\rightarrow\!128$ prediction head maps the sequence back to the original token shape. The editor has approximately 24.9M trainable parameters, including the transformer, prediction head, learned queries, and Qwen LoRA. V-JEPA 2, the renderer bottleneck, Qwen base weights, and the complete renderer remain frozen.

At inference, an MLLM predicts the target caption from the source video and instruction. Qwen-Image-Edit modifies the first frame only when the new dynamics require a changed initial pose or state. The editor prediction, target caption, and resulting first frame then condition the frozen renderer.

\subsection{Counterfactual Data}
\label{app:counterfactual_data}
Figure~\ref{fig:data_pipeline} defines the construction sequence. (1) We filter the same motion-rich internal video collection used for renderer training and assign each clip to an applicable transformation family. (2) An MLLM reasons about a feasible counterfactual outcome and writes the edit instruction, target caption, endpoint states, and motion layout. (3) Qwen-Image-Edit changes the first and/or last frame when the target requires a new pose, object state, or interaction endpoint. (4) VACE renders the counterfactual video from the grounded endpoints, layout, and caption. (5) A VLM quality checker jointly verifies subject identity, unrelated scene context, visual quality, and completion of the requested temporal outcome. (6) Every accepted pair is reused in both directions by swapping source and target and reversing the instruction.

Table~\ref{tab:counterfactual_stats} records 11,702 generated candidates, of which 9,964 pass quality control (85.1\%). After final decoding and cache checks, 9,946 usable pairs remain, yielding approximately 20K directed examples across 17 edit directions. Pass rates range from 64.3\% for break-to-intact to 96.6\% for open-to-close, illustrating why explicit filtering is necessary.

\begin{table*}[t]
\centering
\caption{Counterfactual generation and VLM quality-control statistics.}
\label{tab:counterfactual_stats}
\small
\setlength{\tabcolsep}{3.2pt}
\begin{tabular}{llrrrr}
\toprule
Group & Generated direction & Before & Pass & Filtered & Pass rate \\
\midrule
\multirow{4}{*}{Action}
 & activity $\rightarrow$ sleep & 1,994 & 1,875 & 119 & 94.0\% \\
 & move $\rightarrow$ stop & 480 & 370 & 110 & 77.1\% \\
 & run $\rightarrow$ walk & 3,000 & 2,469 & 531 & 82.3\% \\
 & reduce distance & 2,000 & 1,927 & 73 & 96.4\% \\
\midrule
\multirow{4}{*}{Interaction}
 & break $\rightarrow$ intact & 196 & 126 & 70 & 64.3\% \\
 & cut $\rightarrow$ whole & 3,000 & 2,262 & 738 & 75.4\% \\
 & open $\rightarrow$ close & 266 & 257 & 9 & 96.6\% \\
 & throw $\rightarrow$ hold & 294 & 283 & 11 & 96.3\% \\
\midrule
World & move $\rightarrow$ avoid & 472 & 395 & 77 & 83.7\% \\
\midrule
\multicolumn{2}{l}{\textbf{Total}} & \textbf{11,702} & \textbf{9,964} & \textbf{1,738} & \textbf{85.1\%} \\
\bottomrule
\end{tabular}
\end{table*}

\subsection{Editor Training}
\label{app:editor_training}
Static-to-real pretraining first constructs 169K pseudo pairs by repeating a real video's first frame as a static source and using the real clip token as its target. This exposes the editor to the broad distribution of valid dynamics before task-specific supervision. Counterfactual fine-tuning then teaches source-to-target dynamics changes using the sampled directions above. Source-token noise with standard deviation 0.1 improves robustness. We train all editor parameters globally with a peak learning rate of $2\times10^{-4}$, linear warmup, and cosine decay; the two stages take approximately five days on eight NVIDIA H200 GPUs.

For predicted token $\widehat X$ and target $X^*$, define
\begin{equation}
\mu(X)=\operatorname{mean}_tX,
\end{equation}
\begin{equation}
A(X)=X-\mu(X).
\end{equation}
The objective in Eq.~\ref{eq:loss_v2} combines
\begin{align}
\mathcal L_{\mathrm{dc}} &= \operatorname{MSE}(\mu(\widehat X),\mu(X^*)), \\
\mathcal L_{\mathrm{ac}} &= \operatorname{MSE}(A(\widehat X),A(X^*)), \\
\mathcal L_{\mathrm{cos}} &= 1-\mathbb E_{t,s}[\cos(A(\widehat X)_{t,s,:},A(X^*)_{t,s,:})], \\
\mathcal L_{\mathrm{var}} &= \|\mathbb E_tA(\widehat X)^2-\mathbb E_tA(X^*)^2\|_1, \\
\mathcal L_{\mathrm{fft}} &= \||\operatorname{rFFT}_t(\widehat X)|-|\operatorname{rFFT}_t(X^*)|\|_1.
\end{align}
Their weights are $0.5,1.0,2.0,1.0,$ and $0.5$. The terms separately preserve the temporal mean component, temporal residual, residual direction, motion energy, and rhythm.

\paragraph{Training ablations.}
AC-cos measures residual-direction agreement; AC MSE measures residual magnitude and position; variance error compares temporal energy; and spectrum error compares temporal Fourier magnitude. In Table~\ref{tab:editor_training_ablation}(a), two-stage training improves AC-cos from 0.2629 to 0.2892 and reduces AC MSE from 0.6977 to 0.6352. Panel (b) reports changes relative to its own full-loss reference under the loss-ablation protocol: removing cosine alignment lowers AC-cos by 0.0134, while removing variance and spectrum matching raises their errors by 0.2079 and 0.3276. Reporting deltas avoids conflating the distinct full references used by the training-strategy and loss studies.

\subsection{Dynamics Editing Experiments}
\begin{figure*}[t]
\centering
\includegraphics[width=\textwidth]{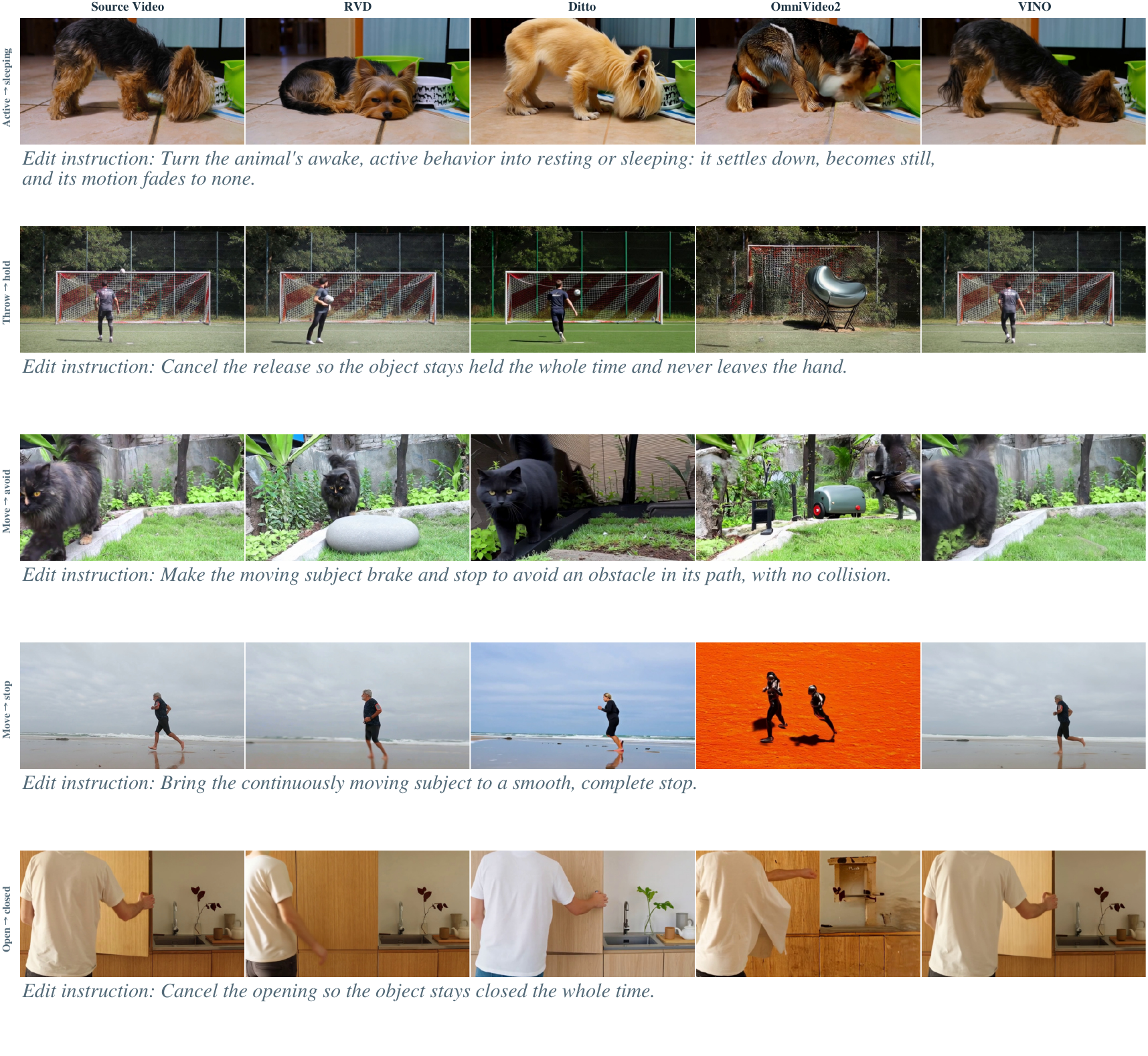}
\caption{\textbf{More editor comparisons.} Five dynamics edits compare RVD with Ditto~\citep{Bai_2025_Ditto}, OmniVideo2~\citep{Yang_2026_OmniVideo2}, and VINO~\citep{Chen_2026_VINO} at aligned relative time; each row reports its instruction.}
\label{fig:editor_baseline_gallery}
\end{figure*}

The editor benchmark covers all 450 candidate public-web cases across 17 directed transformations. Four methods produce 1,800 outputs. IF and Pass measure edit completion. VBench measures background consistency and flicker within each output. Reduced motion can itself improve flicker consistency. RVD achieves the highest IF (4.144), Pass (80.0\%), and temporal flicker consistency (0.9908) (Table~\ref{tab:editor_selected}). Figure~\ref{fig:editor_baseline_gallery} shows additional examples in which RVD changes the event while other instruction-based editors often retain the source motion.

\section{Benchmark and Evaluation Protocol}
\label{app:selected_benchmarks}

\subsection{Renderer Benchmark}
\label{app:renderer_benchmark}
The source videos cover people, animals, vehicles, and objects in indoor and outdoor scenes with varied viewpoints and motions. Edits include foreground replacement, background replacement, and style transfer. The evaluation and renderer training sets have disjoint raw video IDs and file hashes.

We evaluate RVD and six video-editing baselines on the same source video and instruction, with one generated output per method and no reranking. IF is the mean 1--5 instruction-following score; Pass is the percentage judged to complete the edit. JEPA is the cosine similarity between source and output V-JEPA features and serves as a motion-retention proxy that also reflects visual semantics. Imaging is the VBench imaging-quality score~\citep{Huang_2024_VBench}. The fixed-token flow intervention in Table~\ref{tab:flow_intervention} provides an additional encoder-independent motion measurement. Table~\ref{tab:renderer_selected} reports the aggregate comparison, and Figure~\ref{fig:renderer_baseline_gallery} illustrates the three appearance changes.

\subsection{Editor Benchmark}
\label{app:benchmark_protocol}
The editor benchmark contains all 450 candidate dynamics-edit cases from public web videos and 1,800 outputs from RVD, Ditto, OmniVideo2, and VINO. Its 17 directed transformations cover animal activity, locomotion, speed and travel distance, object state, interaction, and causal outcome. The source videos vary in subject, scene, viewpoint, motion magnitude, and temporal pattern. Raw video IDs and file hashes are disjoint from editor training data.

\paragraph{Target construction and method inputs.}
For RVD, Qwen3-VL-32B plans the target caption from eight source frames and the edit instruction. Qwen-Image-Edit changes the first frame when the requested dynamics require a different initial pose or state. Each baseline receives the same source video, including its original first frame, and the same instruction through its native interface. Ditto uses video conditioning without a separate reference image, VINO takes the instruction in its caption field, and OmniVideo2 uses the source clip and edit prompt. Each method produces one sample under a fixed configuration without reranking. Table~\ref{tab:editor_selected} compares these complete pipelines, including RVD's target-context preparation. The within-RVD comparison in Table~\ref{tab:editor_necessity} fixes the target frame and caption to isolate the dynamics input.

\paragraph{Metrics.}
We use Qwen3-VL-8B-Instruct to score instruction following from the edit instruction and 12 uniformly sampled source/output frames, with deterministic decoding. IF is the mean 1--5 score; Pass requires \texttt{performed=true} and score $\geq3$. All methods use the same evaluator. The rubric assigns 5 to full completion, 4 to mostly completed edits, 3 to partial or ambiguous edits, 2 to incorrect attempts, and 1 to absent edits. Pass therefore includes partial edits. Human raters assess the temporal process separately. Six experts assessed the same methods in two panels of three, with each panel assigning joint scores (Appendix~\ref{app:human_questionnaire}). VBench background consistency and temporal flickering measure output-internal stability; source-motion similarity is assessed separately through flow; DINO measures appearance consistency across output frames.

\section{Controlled Evaluation of Dynamics Editing}
\label{sec:editor_necessity}
\label{app:fixed_context_editor}

\textbf{Motivation.}
We assess how dynamics-token manipulation controls event evolution, motion trajectories, and timing. Controlled interventions isolate the dynamics pathway, while edit-family comparisons characterize the flexibility of the shared representation across different transformations.

\textbf{Controlled comparison.}
We evaluate 112 matched cases spanning 17 directed subtasks. For every case, we fix the target first frame, caption, instruction, renderer, sampling schedule, and random seed, and vary only the dynamics input. We compare four settings: no dynamics token, the unedited source token, a matched donor token from the same task, and the edited token predicted from the correct source. This isolates the dynamics-editing pathway within the unified RVD framework. The control manifest uses up to ten cases per subtask in its original order.

\textbf{Editor contribution.}
The edited source token yields the highest within-output appearance consistency of the four conditions (Table~\ref{tab:editor_necessity}). The unedited token often retains the original motion. In the donor examples, both motion and appearance can change. These measurements capture sensitivity to the dynamics input and consistency within the output.

\begin{table}[t]
\centering
\caption{\textbf{Within-output consistency on 112 matched cases with fixed target context.} We vary only the dynamics input. DINO measures appearance consistency across frames of each output.}
\label{tab:editor_necessity}
\small
\setlength{\tabcolsep}{4.5pt}
\begin{tabular}{lr}
\toprule
Dynamics condition & DINO $\uparrow$ \\
\midrule
No dynamics token & 0.864 \\
Unedited source & 0.808 \\
Matched video, same task & 0.826 \\
Edited source (ours) & \textbf{0.921} \\
\bottomrule
\end{tabular}
\end{table}

\textbf{Metric scope.}
We extract DINO ViT-B/16 features from six output frames and average the five cosine similarities between the first frame and each later frame. DINO thus measures appearance consistency within the output. Source-flow similarity measures how much motion structure carries over from the source.

\textbf{Control across edit families.}
Table~\ref{tab:fixed_context_groups} separates trajectory/rhythm edits from state/interaction edits. Trajectory and rhythm tasks include starting, stopping, speed change, distance control, and avoidance. The table reports within-output appearance consistency for each group. These tasks exercise different aspects of temporal control within the same learned dynamics space, complementing the state and interaction edits evaluated alongside them.

The controlled experiment isolates dynamics conditioning within RVD. The 450-case benchmark compares complete systems, including their preparation of the target context.

\begin{table*}[t]
\centering
\caption{Within-output consistency by edit family. Target context and sampling conditions are identical within each case.}
\label{tab:fixed_context_groups}
\small
\setlength{\tabcolsep}{4.5pt}
\begin{tabular}{llr}
\toprule
Edit family & Dynamics condition & DINO $\uparrow$ \\
\midrule
\multirow{4}{*}{Trajectory / rhythm}
 & None & 0.859 \\
 & Unedited source & 0.840 \\
 & Matched video & 0.801 \\
 & Edited source & \textbf{0.912} \\
\midrule
\multirow{4}{*}{State / interaction}
 & None & 0.868 \\
 & Unedited source & 0.781 \\
 & Matched video & 0.846 \\
 & Edited source & \textbf{0.928} \\
\bottomrule
\end{tabular}
\end{table*}

\textbf{Pairwise source-motion comparison.}
We further compare the edited source token with a matched donor token using a pairwise video judge. The judge prefers the edited source token in 65.2\% of cases, compared with 33.0\% for the donor token and 1.8\% ties. This comparison favors edited tokens for retaining source-motion characteristics while completing the requested edit. Source-flow similarity measures how much of the original motion remains. Higher similarity is desirable when the instruction calls for preserving that motion.

\section{Human Evaluation: Questionnaire and Results}
\label{app:human_questionnaire}
\textbf{Scope.}
Six experts participated in a blinded evaluation. Two panels of three each assigned one joint score per dimension to every candidate. Across 65 cases and 260 candidate videos, this yielded 520 panel--candidate ratings. We average the two panel scores. Each panel completed all three dimensions for every candidate. We report agreement between the panels below.

\textbf{Blinded presentation.}
The evaluation contains 34 complete-system comparisons and 31 fixed-condition comparisons, each with four anonymous candidates. Cases were ordered by a fixed seeded hash within each task, with at most two cases per task; existing scores were not used to choose cases. Candidate identities and model scores were hidden. The form instructed each panel to watch complete videos at their native playback rates, with replay and pausing permitted. The two comparison protocols are analyzed separately.

\subsection{Questionnaire Shown to Raters}
\textbf{Record fields:} Anonymous panel ID; case ID; candidate label; edit instruction. Method names are not displayed.

\textbf{Viewing instructions.}
Watch the full source video and each candidate before scoring. Evaluate each candidate independently. A correct final state does not imply a correct event process. Do not penalize a change that the instruction explicitly requests. If the evidence is insufficient, select ``cannot assess'' and explain why. Do not infer method identity or infer unobserved temporal behavior from a single frame.

Each question uses the same response scale: \textbf{1: not met; 2: partially met; 3: fully met; or cannot assess}.
\begin{enumerate}
\item \textbf{Requested-event completion.} Does the candidate perform the action or event requested by the instruction?
\item \textbf{Temporal-process correctness.} Is the action or event process correct throughout the video, including action order, transitions, and contact? Judge the process, not only the final state. Do not invent a precise trajectory or timestamp absent from the instruction.
\item \textbf{Preservation of unrelated content.} Are the identities, scene, and relationships that should remain unchanged preserved? Required changes in pose, speed, or state are not preservation errors.
\end{enumerate}
An optional comment records an error timestamp, a visible issue, or the reason a dimension cannot be assessed.

The offline form saves progress separately for each anonymous panel ID. Re-entering an ID resumes its record; a new ID starts an independent record. Panels export their joint ratings as JSON upon completion. We analyze ratings from the final three-point protocol separately from the earlier five-point draft.

\subsection{Complete-System Results and Qwen Comparison}
Table~\ref{tab:human_system} compares human scores on a three-point scale and Qwen IF scores on a five-point scale for the same 34 cases. RVD has the highest human mean on every dimension, both overall and within each panel. Mean human completion and Qwen IF rank the methods in the same order, from RVD to VINO, Ditto, and OmniVideo2. This agreement concerns the aggregate ranking.

\begin{table*}[t]
\centering
\caption{\textbf{Human evaluation of complete systems: six experts in two panels, 34 cases per method.} Human dimensions use 1--3; Qwen IF uses 1--5. Strict process success requires both human completion and process scores to equal three. Qwen Pass uses the automated rubric.}
\label{tab:human_system}
\small
\begin{tabular}{lrrrrr}
\toprule
Method & Completion & Process & Preservation & Strict process (\%) & Qwen IF \\
\midrule
RVD & \textbf{2.029} & \textbf{1.941} & \textbf{1.941} & \textbf{13.2} & \textbf{3.941} \\
VINO & 1.559 & 1.544 & 1.676 & 4.4 & 3.294 \\
Ditto & 1.206 & 1.265 & 1.353 & 1.5 & 2.382 \\
OmniVideo2 & 1.015 & 1.044 & 1.015 & 0.0 & 2.294 \\
\bottomrule
\end{tabular}
\end{table*}

Strict process success is computed separately for each panel--candidate record before averaging: both completion and process must equal three in that record. Thus each method has 68 panel--case records in the complete-system comparison and 62 in the control comparison; each panel contributes repeated ratings.

Each dynamics task contributes two cases, so the complete-system case means also equal task-macro means. Relative to VINO, RVD's paired mean differences are $+0.471$ for completion, $+0.397$ for process correctness, and $+0.265$ for preservation. The corresponding percentile bootstrap 95\% intervals are $[0.162,0.765]$, $[0.132,0.647]$, and $[0.000,0.529]$. We first average the two panels within each case and method, then resample paired cases 20,000 times with seed 20261115. We keep both panel ratings together when resampling cases and report intervals without a multiple-comparison adjustment. RVD has the highest mean preservation score among the methods, and its paired interval against VINO touches zero.

\textbf{Agreement between panels.}
Across the 260 common candidate videos, exact agreement is 63.8\%, 68.1\%, and 64.2\% for completion, process, and preservation, respectively. Linear-weighted Cohen's $\kappa$ is 0.466, 0.487, and 0.443 for the same dimensions. The second panel uses the highest score less frequently, so absolute ratings differ even though both panels rank RVD first on all three complete-system means. These agreement statistics pool both comparison protocols and describe agreement between these two panels.

\subsection{Fixed-Condition Human Results}
Table~\ref{tab:human_control} retains all four conditions on the 31 rated control cases. Edited dynamics outperform unedited and donor dynamics in the reported means. Relative to no dynamics, the mean differences are $-0.016$ in completion, $+0.048$ in process, and $0.000$ in preservation. All three paired bootstrap intervals include zero. Strict process success is equal for edited and no-dynamics conditions.

\begin{table*}[t]
\centering
\caption{\textbf{Fixed-condition human evaluation: two panels, 31 cases.} C: completion; P: process correctness; R: unrelated-content preservation. The macro columns weight the 17 tasks equally. All values use the three-point scale.}
\label{tab:human_control}
\small
\begin{tabular}{lrrrrrr}
\toprule
& \multicolumn{3}{c}{Case mean} & \multicolumn{3}{c}{Task-macro mean} \\
\cmidrule(lr){2-4}\cmidrule(lr){5-7}
Dynamics condition & C & P & R & C & P & R \\
\midrule
No dynamics & \textbf{1.919} & 1.839 & \textbf{1.919} & \textbf{1.971} & 1.868 & \textbf{1.941} \\
Unedited source & 1.355 & 1.339 & 1.500 & 1.382 & 1.397 & 1.544 \\
Matched donor & 1.565 & 1.484 & 1.435 & 1.588 & 1.500 & 1.456 \\
Edited source & 1.903 & \textbf{1.887} & \textbf{1.919} & 1.926 & \textbf{1.912} & 1.926 \\
\bottomrule
\end{tabular}
\end{table*}

\textbf{Interpretation.}
We fixed the questionnaire before analyzing responses and retained every candidate. The means summarize ordinal ratings. The strict process criterion captures errors that a completion score alone can miss. Qwen Pass uses a different scale and completion threshold.

\subsection{Illustrative Cases of Token-Editing Benefits}
\label{app:token_benefit_examples}
\begin{figure*}[t]
\centering
\includegraphics[width=\textwidth]{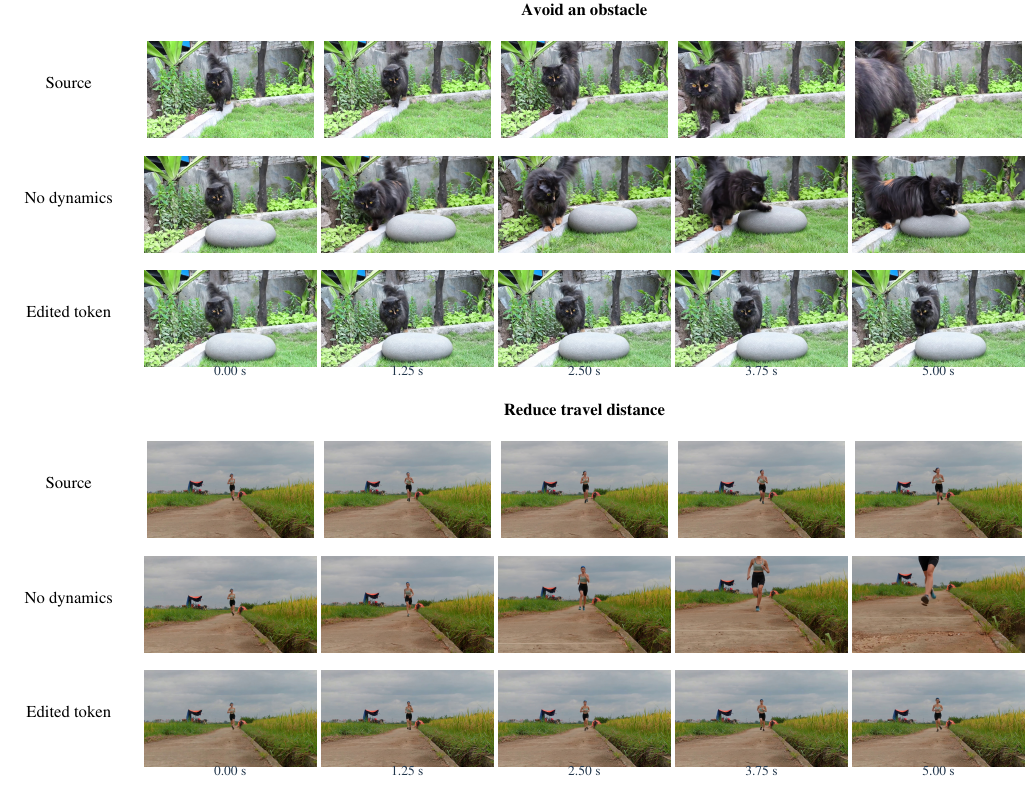}
\caption{\textbf{Selected examples illustrating token-editing benefits under fixed target conditions.} Top: obstacle avoidance. The no-dynamics output advances onto the obstacle, whereas the edited-token output keeps the cat behind it. Bottom: reduced travel distance. The no-dynamics runner approaches and reaches the foreground, whereas the edited-token runner remains farther back. The columns show 0\%, 25\%, 50\%, 75\%, and 100\% of each clip's frame-index range; displayed times refer to the generated outputs. Original frames are shown without retouching.}
\label{fig:token_benefits}
\end{figure*}

Figure~\ref{fig:token_benefits} presents selected positive examples from the existing fixed-condition human evaluation. The source, no-dynamics output, and edited-token output are shown at aligned relative positions. Target frame, caption, renderer, and sampling are held fixed between the two output conditions. These examples contain 81-frame source clips at 24 fps and 81-frame output clips encoded at 16 fps, as verified from the media files. Columns align clips by relative frame index. The last source frame is at 3.33 s and the last output frame at 5.00 s. This encoding difference is separate from the token-resampling operations in Appendix~\ref{app:retime}.

\paragraph{Motion progression and displacement.}
The avoidance example illustrates a difference in progression relative to an obstacle. With no dynamics, the cat advances onto it; with edited tokens, the cat remains behind it. The distance example instead illustrates the extent of travel: the no-dynamics runner reaches the foreground, while the edited-token runner remains farther from the camera. These comparisons visualize two distinct temporal outcomes under the same target conditions.

\begin{figure*}[t]
\centering
\includegraphics[width=\textwidth]{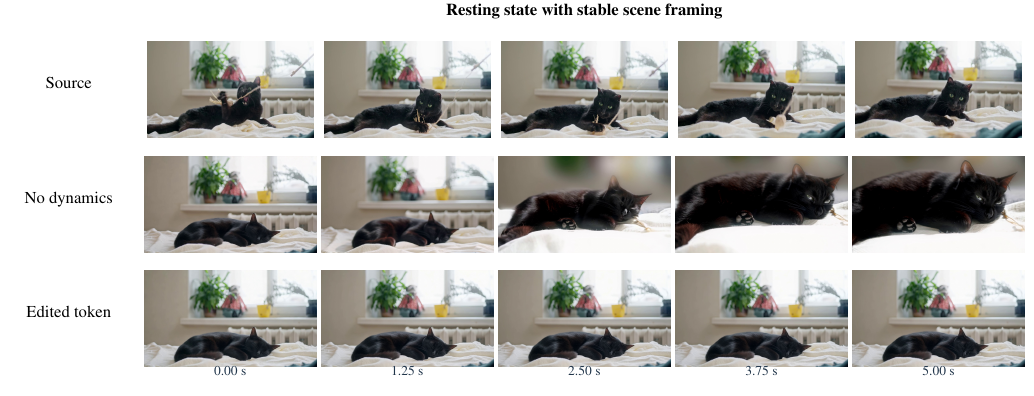}
\caption{\textbf{Selected example of scene stability during an activity-state edit.} The source shows an active cat. Both output conditions begin in a resting pose, but the no-dynamics output changes to a close-up, whereas the edited-token output retains the initial scene framing across the displayed times. This state-replacement example shows how framing changes after the resting pose is established. Frames use the same relative positions as Fig.~\ref{fig:token_benefits}.}
\label{fig:token_benefits_state}
\end{figure*}

\paragraph{Preserving scene context during a state edit.}
Figure~\ref{fig:token_benefits_state} highlights a complementary benefit: maintaining the broader scene while changing the activity state. The edited-token sequence keeps the bed, window, and subject scale consistent across the displayed frames. Together, these selected comparisons make the behavior of the shared dynamics interface visible through motion progression, displacement, and scene stability. They supplement the aggregate evaluation in Appendix~\ref{app:human_questionnaire}.

\section{Evidence for Editing and Reusing the Dynamics Representation}
\label{app:evidence_synthesis}
We revisit the reconstruction, editing, and human evaluation results to ask three questions. How much source motion can the token reproduce? How does editing the source token compare with keeping it or using a donor? What does the edited token add when the target context is fixed? We use the existing runs, videos, and ratings throughout.

\subsection{What Is Gained by Editing the Source Tokens?}
The fixed-condition protocol holds target frame, caption, renderer, and sampling fixed. Its three comparisons address different alternatives. Unedited source tokens test whether the original dynamics can simply be reused for a changed event. Same-task donor tokens test whether a generic example of the requested dynamics can replace a source-conditioned transformation. No dynamics tests whether the visual and textual target conditions suffice without that pathway.

Table~\ref{tab:editor_paired_evidence} reports paired differences from the human evaluation. Edited source tokens score higher than both unedited source tokens and same-task donors on completion, process correctness, and preservation. The paired intervals for all six differences lie above zero.

\begin{table*}[t]
\centering
\caption{\textbf{Incremental effects under identical target conditions.} Edited-source score minus each alternative, on the same 31 human-rated control cases. Each cell gives mean difference [95\% paired case-bootstrap interval]. C, P, R denote completion, process correctness, and preservation. All alternatives are reported.}
\label{tab:editor_paired_evidence}
\small
\begin{tabular}{lccc}
\toprule
Alternative & C & P & R \\
\midrule
Unedited source & $+0.548$ $[0.274,0.806]$ & $+0.548$ $[0.306,0.774]$ & $+0.419$ $[0.161,0.661]$ \\
Same-task donor & $+0.339$ $[0.097,0.597]$ & $+0.403$ $[0.161,0.645]$ & $+0.484$ $[0.242,0.726]$ \\
No dynamics & $-0.016$ $[-0.275,0.242]$ & $+0.048$ $[-0.194,0.290]$ & $+0.000$ $[-0.226,0.226]$ \\
\bottomrule
\end{tabular}
\end{table*}

We average the two panel scores within each case, then resample paired cases 20,000 times (seed 20261115). The intervals apply to these panels and cases and use no adjustment for multiple comparisons. Each three-person panel contributed one joint judgment per candidate. Both analyses use the same ratings.

\subsection{Variation Across Task Directions}
Table~\ref{tab:editor_all_tasks} shows the difference between edited tokens and no dynamics for every rated task direction. Each mean summarizes one or two cases per direction. The instruction specifies whether the edit calls for a visible transition or a replacement state.

\begin{table*}[t]
\centering
\caption{\textbf{All task directions in the existing human control records.} Differences are edited source minus no dynamics after averaging the two panel scores within each case. Positive values favor editing. Each mean summarizes the rated cases for one task.}
\label{tab:editor_all_tasks}
\small
\begin{tabular}{lrrrr}
\toprule
Task direction & Cases & $\Delta$ completion & $\Delta$ process & $\Delta$ preservation \\
\midrule
animal\_activity2sleep & 2 & $+0.500$ & $+0.750$ & $+0.750$ \\
animal\_sleep2activity & 2 & $+0.250$ & $+0.000$ & $+0.250$ \\
break2intact & 2 & $+0.250$ & $+0.500$ & $+0.250$ \\
close2open & 2 & $-0.500$ & $-0.250$ & $-0.250$ \\
cut2whole & 2 & $+0.500$ & $+0.000$ & $-0.250$ \\
extend\_distance & 1 & $-0.500$ & $+0.000$ & $+0.000$ \\
hold2throw & 2 & $+0.250$ & $+0.250$ & $+0.000$ \\
intact2break & 1 & $-0.500$ & $+0.000$ & $+0.000$ \\
move2avoid & 2 & $+1.250$ & $+1.250$ & $+1.000$ \\
move2stop & 2 & $+0.000$ & $+0.000$ & $+0.250$ \\
open2close & 2 & $-0.250$ & $-0.250$ & $-0.250$ \\
reduce\_distance & 2 & $+1.000$ & $+0.750$ & $+0.500$ \\
run2walk & 2 & $+0.000$ & $+0.000$ & $+0.250$ \\
stop2move & 2 & $-0.750$ & $-0.500$ & $-0.500$ \\
throw2hold & 2 & $-1.000$ & $-0.750$ & $-0.750$ \\
walk2run & 2 & $-1.250$ & $-1.000$ & $-1.000$ \\
whole2cut & 1 & $+0.000$ & $+0.000$ & $-0.500$ \\
\bottomrule
\end{tabular}
\end{table*}

\subsection{Complementary Evidence for the Shared Representation}
Table~\ref{tab:representation_evidence} summarizes three properties of the shared representation. Removing the token reduces reconstruction fidelity. Keeping it fixed across appearance changes preserves measurable correspondence with source motion. Resampling its temporal axis changes motion pace without additional training.

\begin{table*}[t]
\centering
\caption{\textbf{Properties of the dynamics representation measured by each experiment.} The paired human comparisons quantify the editor's effect on event scores.}
\label{tab:representation_evidence}
\small
\begin{tabular}{p{.20\textwidth}p{.40\textwidth}p{.32\textwidth}}
\toprule
Question & Existing evidence & Supported interpretation \\
\midrule
Renderable source information & Full/no-dynamics PSNR: 19.261/16.663; SSIM: 0.5761/0.5216 (Table~\ref{tab:renderer_arch_main}). & The dynamics pathway contributes to reconstruction. \\
Spatially organized injection & Full/no-RoPE PSNR: 19.261/18.222; no-centering: 18.427 (Table~\ref{tab:renderer_arch_main}). & Spatial alignment and centering help the tested reconstruction setting. \\
Motion reuse under appearance change & Matched/mismatched flow trajectory: 0.727/0.312; motion profile: 0.501/0.012 (Table~\ref{tab:flow_intervention}). & Output motion aligns more closely with the matched source. \\
Direct temporal operations & Explicit resampling of the temporal axis with the same renderer (Appendix~\ref{app:retime}). & Resampling controls pace without extra training. \\
Transforming source dynamics & Positive human differences versus unedited source and donor tokens (Table~\ref{tab:editor_paired_evidence}). & Editing improves these alternatives under fixed target conditions. \\
\bottomrule
\end{tabular}
\end{table*}

\paragraph{Training reuse and efficiency.}
All three operations reuse the renderer. Retiming and appearance changes require no additional task-specific training. Semantic editing uses approximately 20K directed examples after motion pretraining on 169K pseudo-pairs and initialization from pretrained models. Rank and entropy describe variation within the learned representation. The editing comparisons measure the benefit of transforming a source token relative to keeping it or replacing it with a donor.

\subsection{Source Context as an Editable and Reusable Input}
\label{app:token_reuse}
For each source clip $V_s$, the encoder extracts a dynamics token $D_s=E(V_s)$. The renderer takes the result of applying an operation to this token,
\begin{equation}
\widehat V=R_\theta(I,c,\mathcal O(D_s)).
\end{equation}
For semantic editing, $\mathcal O$ is the instruction-conditioned editor with the inputs in Eq.~\ref{eq:editor_token}. For retiming, it is a temporal index map. For appearance changes, it is the identity. All three operations use RVD's trained renderer.

\paragraph{Where source context enters the comparison.}
In Figs.~\ref{fig:token_benefits} and~\ref{fig:token_benefits_state}, the edited-token and no-token outputs share the same target frame and caption. Varying the dynamics input changes how far a subject travels, where it moves relative to an obstacle, and how the scene is framed. Across rated cases, editing the source token improves on keeping it or substituting a donor (Table~\ref{tab:editor_paired_evidence}).

\paragraph{Passing dynamics to appearance-editing tasks.}
For foreground replacement, background replacement, and style changes, RVD passes $D_s$ unchanged to the same renderer while replacing $(I,c)$. The source-derived representation thus provides the shared motion input across these visual modifications. Figure~\ref{fig:rerender_examples} illustrates all three, and Table~\ref{tab:flow_intervention} measures correspondence to the original motion independently of the V-JEPA encoder. Reusing the source token preserves the event across these appearance changes.

\begin{table*}[t]
\centering
\caption{\textbf{One source-derived representation, multiple editing operations.} The same trained renderer is reused in every row. Additional training refers to training after the shared renderer and representation have been learned.}
\label{tab:token_operations}
\small
\begin{tabular}{p{.23\textwidth}p{.29\textwidth}p{.38\textwidth}}
\toprule
Request & Operation on source token & Additional task training / evidence \\
\midrule
Change an action or interaction & Predict edited token from $D_s$ and semantic conditions & Learned editor; complete-system and fixed-condition comparisons. \\
Adjust motion pace & Resample the temporal positions of $D_s$ & None; direct token manipulation with the frozen renderer. \\
Change foreground or background & Keep $D_s$; replace visual conditions & None for RVD dynamics/rendering modules; image preparation uses an existing image editor. \\
Change visual style & Keep $D_s$; replace visual conditions & None for RVD dynamics/rendering modules; motion reuse is examined through the flow intervention. \\
\bottomrule
\end{tabular}
\end{table*}

\subsection{Learning with a Limited Counterfactual Update Budget}
\label{app:limited_budget}
Table~\ref{tab:editor_training_ablation}(a) compares two initializations on the same \texttt{intact2break} split of 118 training pairs and eight validation pairs. Both runs use the same frozen foundation models and renderer bottleneck, input features, normalization, losses, optimizer, batch size, seed, and number of counterfactual updates. We initialize the editor and semantic queries either from scratch or from the static-to-real checkpoint at step 12,000. Both runs operate in token space.

Each arm receives 100 counterfactual updates at batch size one, learning rate $2\times10^{-4}$, seed 42, and 3\% warmup. The existing Qwen LoRA unfreeze schedule starts at step 300, so LoRA remains frozen during this diagnostic. Evaluation uses the final step-100 checkpoint and averages the same eight validation pairs equally.

\begin{table}[t]
\centering
\caption{\textbf{Editor initialization under a fixed update budget.} Same 118-pair training pool and 100 updates per arm; eight validation pairs. AC measures temporal residuals; DC measures the temporal mean.}
\label{tab:limited_budget_tokens}
\small
\begin{tabular}{lrrr}
\toprule
Initialization & AC-cos $\uparrow$ & AC MSE $\downarrow$ & DC MSE $\downarrow$ \\
\midrule
Fresh editor & 0.2629 & 0.6977 & \textbf{0.2668} \\
Static-to-real & \textbf{0.2892} & \textbf{0.6352} & 0.3032 \\
\bottomrule
\end{tabular}
\end{table}

At the same update budget, static-to-real initialization improves residual-direction agreement by 0.0263 and reduces AC MSE by 0.0625. Copying the source gives a reference AC-cos of 0.2784. DC MSE increases by 0.0364.

We use one task and one seed. Total pretraining compute differs between runs. We excluded the validation pairs from both fine-tuning runs. We did not check those pairs against the earlier static-to-real corpus. The full training budget includes approximately 169K renderer clips, 169K pseudo-pairs, and approximately 20K directed counterfactual examples, alongside pretrained foundation models.

\paragraph{What the no-token control can establish.}
The no-token control removes the dynamics input from the trained renderer at inference. It measures the effect on rendered events under fixed target conditions. Table~\ref{tab:limited_budget_tokens} compares editor initializations after 100 counterfactual updates.

\subsection{Training Data and Reuse Across Editing Tasks}
\label{app:training_data_comparison}
Table~\ref{tab:baseline_training_data} places RVD's training recipe alongside the original reports for the baselines evaluated in this paper. The useful distinction is between learning a reusable video interface and collecting paired supervision for each edit. RVD learns reconstruction and motion priors before fitting its dynamics editor with 19,892 directed counterfactual examples. The resulting source-derived token also supports appearance changes and temporal resampling through the same renderer, without a separately trained RVD module for either operation.

\begin{table*}[t]
\centering
\caption{\textbf{Reported training data for the evaluated editing systems.} A and D denote the appearance and dynamics tasks evaluated here. Counts cover training after foundation-model pretraining. Each row uses the units reported by the original authors.}
\label{tab:baseline_training_data}
\small
\setlength{\tabcolsep}{4pt}
\begin{tabular}{p{.17\textwidth}p{.05\textwidth}p{.32\textwidth}p{.39\textwidth}}
\toprule
Method & Tasks & Reported data & Scope and source \\
\midrule
InsV2V~\citep{Cheng_2024_InsV2V} & A & 68,410 successful synthetic generations & Sum of 33,421 and 34,989 in original Sec.~3.2; counts include only retained pairs. \\
Ditto / Editto~\citep{Bai_2025_Ditto} & A, D & $>1$M video-edit triplets & Ditto-1M source/instruction/edited-video data; original Secs.~3 and~5.1. \\
Omni-Video 2~\citep{Yang_2026_OmniVideo2} & A, D & $>1$M mixed training instances & Image/video generation and editing combined; original Sec.~3. \\
VInO~\citep{Chen_2026_VINO} & A, D & $\mathcal{O}(100)\mathrm{k}$ samples per stage & Three stages, original Table~1 and Sec.~3.1; image/video mixture. Cross-stage unique-video total unspecified. \\
STDF~\citep{Yatim_2024_STDF} & A & 0 additional task-training examples & Zero-shot motion transfer using pretrained diffusion features. \\
VideoGrain~\citep{Yang_2025_VideoGrain} & A & 0 additional task-training examples & Training-free attention modulation; original Appendix~K. \\
\midrule
\textbf{RVD (ours)} & A, D & $\sim$169K real clips; 169K static-to-real pseudo-pairs; \textbf{19,892 directed edits} & Renderer, editor pretraining, and editor fine-tuning, respectively. Directed edits reuse 9,946 pairs in both directions. The real clips also supply the pseudo-pairs, so the counts overlap. \\
\bottomrule
\end{tabular}
\end{table*}

\paragraph{Limited paired editing supervision, shared task support.}
RVD's roughly 20K directed editing examples are substantially fewer than the million-scale paired edit corpus reported for Ditto. This comparison concerns paired editing supervision: RVD additionally uses the real-video reconstruction and pseudo-pair stages listed above, as well as pretrained visual, language, image-editing, and video-generation components. Appearance changes reuse the source token, and retiming changes its temporal indices; neither requires new paired video-edit training for the RVD renderer. All three operations share the trained renderer.

\paragraph{Relationship to the measured results.}
RVD achieves higher instruction-following scores and pass rates than Ditto, OmniVideo2, and VINO (Table~\ref{tab:editor_selected}). Table~\ref{tab:renderer_selected} reports appearance-editing results, and Table~\ref{tab:flow_intervention} measures motion reuse. These results describe what the shared renderer and an editor trained with limited paired supervision achieve together. STDF and VideoGrain use no additional task-training data. Omni-Video 2 and VInO report combined image and video counts. Table~\ref{tab:limited_budget_tokens} compares initializations within RVD at a fixed update budget.

\section{Limitations}
\label{app:limitations}
RVD edits dynamics in a shared continuous space. The editor learns from sampled counterfactual transformations and predicts one deterministic target trajectory per instruction. This limits open-ended edits for which several motions are valid. The benchmark covers common dynamics families. Future work could combine stochastic dynamics generation with supervision on a wider range of transformations and their compositions.

\paragraph{Task semantics and training scope.}
A state-replacement instruction may allow the clip to begin in the target state. A transition instruction requires the change itself to be visible. This distinction determines whether an edited first frame satisfies the request. The questionnaire scores completion and temporal-process correctness separately, with results pooled across semantic categories. Fine-tuning on approximately 20K directed examples builds on pretrained models and 169K pseudo-pairs.

\paragraph{Failure modes.}
We observe three especially difficult regimes. First, rapid motion can be temporally aliased by the compressed token and rendered with blur, reduced displacement, or a missed intermediate state. The problem is most visible when the requested event occurs between the uniformly sampled temporal positions. Second, the token may encode strong camera motion together with object motion. As a result, editing or transferring it can perturb camera trajectory, destabilize the background, or preserve the wrong egomotion; optical-flow measurements are also confounded in this setting. Third, complex interactions with several independently moving subjects, persistent occlusion, or rapid contact and topology changes can produce identity merging, incorrect contact order, or a plausible endpoint reached through the wrong process. The current training distribution, which emphasizes one dominant subject or interaction, does not fully cover these cases.

\section{Ethics and Reproducibility}
Dynamics editing can alter the apparent causal content of footage. Transparent labeling and provenance safeguards are important for any release. Architecture, training, data construction, and evaluation protocols are described above. Portions of the project depend on internal infrastructure and data; public release of code, data, and model weights requires institutional review.

\end{document}